\pdfoutput=1
\documentclass[11pt]{article}

\usepackage[final]{acl}

\usepackage{times}
\usepackage{latexsym}

\usepackage{booktabs}

\usepackage[T1]{fontenc}
\usepackage[utf8]{inputenc}

\usepackage{microtype}

\usepackage{inconsolata}

\usepackage{graphicx, multirow}
\usepackage{todonotes}
\usepackage{enumitem}
\usepackage{subcaption}

\usepackage{tabularx}
\usepackage{array}
\usepackage{makecell}
\usepackage{float}

\usepackage{amsfonts}

\usepackage{threeparttable}

\newcommand{\App}{OmniTemp}

\title{Beyond Pairwise: Global Zero-shot Temporal Graph Generation}

\author{
    Alon Eirew$^{1}$ \quad Kfir Bar$^{2}$ \quad Ido Dagan$^{1}$ \\
    $^{1}$Computer Science Department, Bar-Ilan University \\
    $^{2}$Efi Arazi School of Computer Science, Reichman University \\
    \texttt{alon.eirew@gmail.com}, \quad \texttt{kfir.bar@runi.ac.il}
}

\begin{document}
\maketitle

\begin{abstract}

Temporal relation extraction (TRE) is a fundamental task in natural language processing (NLP) that involves identifying the temporal relationships between events in a document. 
Despite the advances in large language models (LLMs), their application to TRE remains limited. Most existing approaches rely on \textit{pairwise} classification, where event pairs are classified in isolation, leading to computational inefficiency and a lack of global consistency in the resulting temporal graph. In this work, we propose a novel \textit{zero-shot} method for TRE that generates a document’s complete temporal graph in a single step, followed by temporal constraint optimization to refine predictions and enforce temporal consistency across relations. 
Additionally, we introduce \textit{OmniTemp}, a new dataset with complete annotations for all pairs of targeted events within a document. Through experiments and analyses, we demonstrate that our method outperforms existing zero-shot approaches and offers a competitive alternative to supervised TRE models.

\end{abstract}

\section{Introduction}
\label{intro}

Temporal relation extraction (TRE) is a foundational task in natural language processing (NLP) that supports applications such as event forecasting \cite{Ma2023ContextawareEF}, misinformation detection \cite{lei-huang-2023-identifying}, and medical treatment timeline construction \cite{yao-etal-2024-overview}.

The TRE task is formulated as follows: given a pair of event mentions, identify the temporal relation between them (e.g., before, after, equal, include, is included, vague). The task has seen significant progress in recent years with the development of supervised models \cite{tan-etal-2023-event, niu-etal-2024-contempo}. However, these models require large amounts of training data, which is scarce in most domains and languages, and difficult to obtain due to the complexity of manually annotating such relations \cite{pustejovsky-stubbs-2011-increasing}.

Recent advances in large language models (LLMs) have shown strong capabilities in capturing linguistic patterns \cite{NEURIPS2020_1457c0d6}, performing multi-step reasoning \cite{10.5555/3600270.3602070}, and applying temporal commonsense knowledge \cite{jain-etal-2023-language-models}, positioning them as promising tools to address data scarcity through zero-shot learning \cite{10.5555/3600270.3601883}. However, existing zero-shot LLM-based TRE work has focused on pairwise classification \cite{yuan-etal-2023-zero, chan-etal-2024-exploring}. Pairwise methods face significant computational challenges, particularly in real-world scenarios where the goal is to construct a complete timeline of events from a document. In such cases, all event pairs must be classified, resulting in \( O(n^2) \) inference calls for \( n \) events. This quadratic complexity becomes impractical when using LLMs due to their high computational cost per query. Moreover, because pairwise approaches consider each event pair in isolation, they fail to capture the global temporal structure of the document, often leading to inconsistent or contradictory temporal graphs \cite{wang-etal-2020-joint}. As a result of these challenges, zero-shot applications of LLMs to TRE have largely been regarded as ineffective \cite{wei-etal-2024-llms, niu-etal-2024-contempo, chan-etal-2024-exploring, ning-etal-2024-temporal}.

In contrast, \textit{global} TRE involves predicting the complete set of temporal relations between all event pairs in a document, resulting in a \textit{temporal graph} that captures the holistic temporal structure. This approach enables models to enforce global consistency and jointly reason about relations, essential for accurate temporal understanding. In doing so, it also provides a more scalable alternative to the computational inefficiencies of pairwise modeling.

\begin{figure*}[t!]
\centering
\includegraphics[width=\textwidth]{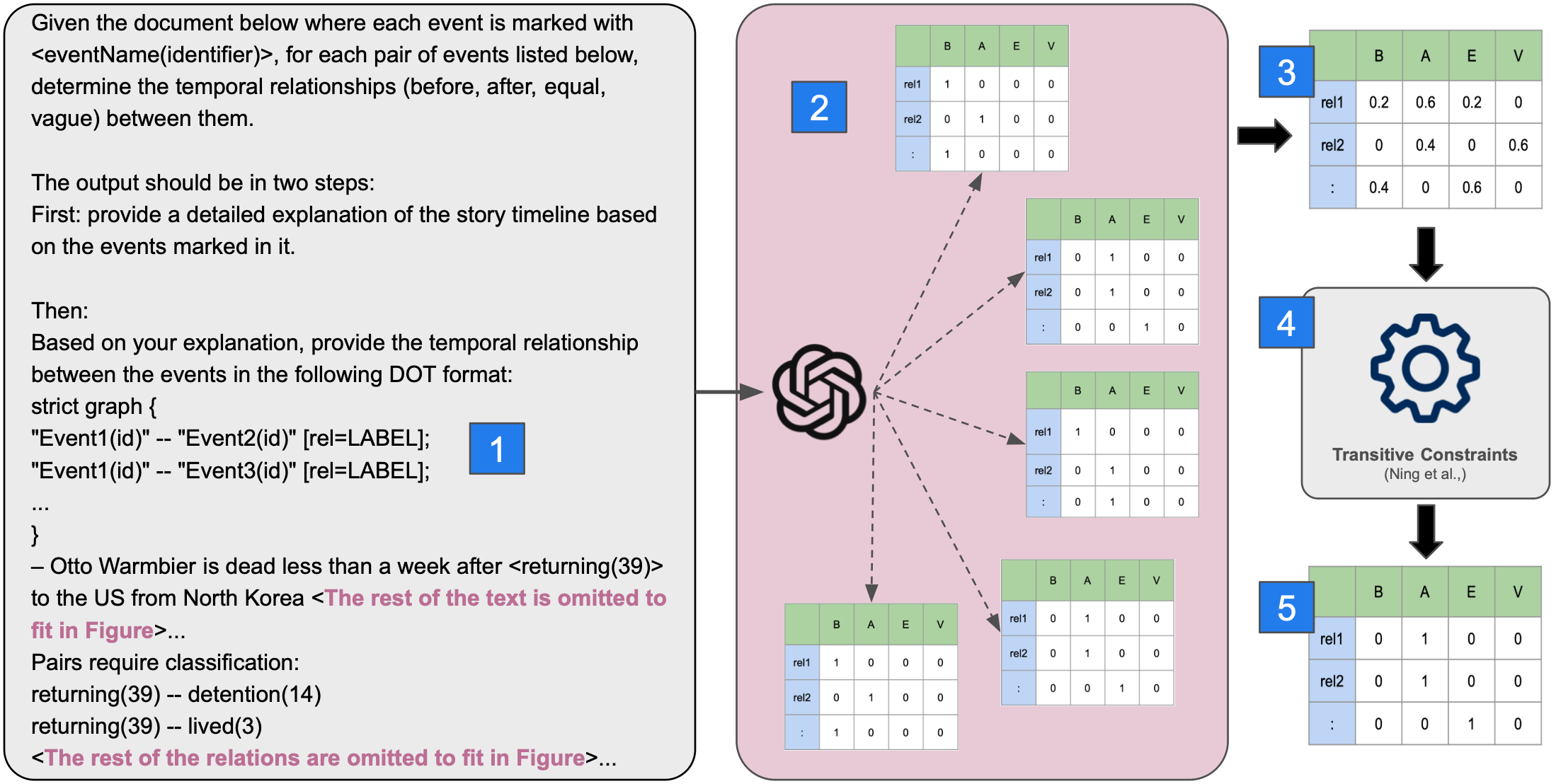}
\caption{Illustration of the pipeline approach (§\ref{section:model}):  
\textbf{[1]} We send the same prompt to the model to generate separate instances of the document's \textit{complete} temporal graph.  
\textbf{[2]} We extract the relation distribution as one-hot vectors over the temporal classes for each relation in each generation.  
\textbf{[3]} We sum and normalize the predictions into a single vector representing the joint prediction over the document's temporal graph.  
\textbf{[4]} We apply an ILP optimization algorithm to this vector.  
\textbf{[5]} The final temporal graph is obtained.
}
\label{fig:figure1}
\end{figure*}

A critical obstacle for global TRE research is the lack of datasets with \textit{complete} temporal relation annotations for all event pairs. Manual annotation of full temporal graphs is notoriously challenging and traditionally considered infeasible \cite{naik-etal-2019-tddiscourse}. Most TRE datasets therefore provide labels only for a subset of event pairs, often limited to events in consecutive sentences \cite{chambers-etal-2014-dense, ning-etal-2018-multi}. This partial coverage constrains models to pairwise strategies and complicates evaluation of long-range temporal reasoning. Alternative automated labeling approaches \cite{naik-etal-2019-tddiscourse, alsayyahi-batista-navarro-2023-timeline} mitigate annotation costs but risk introducing biases inherent to the automated annotation methods themselves.

To address the scarcity of fully annotated datasets, and the inefficiency and global inconsistency of pairwise classification in zero-shot settings, we make the following contributions:\footnote{\App{} and all experimental code are publicly available at \url{https://github.com/AlonEirew/GlobalZeroShotTRE}.}
\begin{itemize}
    \item We propose a novel zero-shot LLM method that generates the entire temporal graph in a single inference step. Our method prompts the model to produce a free-form summary of the event timeline to guide reasoning or ``thinking'', followed by classification of all event pairs, aggregated via a global temporal constraints optimization algorithm to ensure consistency (Figure~\ref{fig:figure1}, and §\ref{section:model}). This approach significantly reduces computational cost compared to pairwise methods while achieving performance competitive with supervised models (§\ref{section:results}).
    \item To support global temporal graph extraction, we introduce \texttt{\App{}}, a new dataset with exhaustive temporal relation annotations for all event pairs (§\ref{section:dataset}).
    \item We discuss how annotation scope and guideline inconsistencies affect zero-shot model assessment. We show that limiting annotations to short-distance event pairs, as well as discrepancies between widely used datasets such as MATRES and TB-Dense, can hinder fair and reliable evaluation of TRE models in zero-shot settings.
\end{itemize}

\section{Background}
\label{section:background}
This section provides relevant background on datasets and zero-shot methods for the temporal relation extraction task.

\subsection{Temporal Relation Extraction Datasets}
\label{section:background:datasets}

The temporal relation extraction task aims to determine the temporal order between pre-extracted events in a text \cite{pustejovsky2003timeml}. For fair and unbiased model evaluation, datasets should provide gold labels for all event pairs or, at a minimum, be randomly sampled from the full set. However, most existing datasets for temporal relation extraction provide only partial annotation due to the complexity and cost of the process \cite{pustejovsky-stubbs-2011-increasing, naik-etal-2019-tddiscourse}. As a result, the two most widely used datasets, MATRES \cite{ning-etal-2018-multi} and TimeBank-Dense (TB-Dense) \cite{chambers-etal-2014-dense}, annotate only relations between events in consecutive sentences.

Recently, the NarrativeTime project \cite{rogers-etal-2024-narrativetime}, a large effort of expert annotation, released a comprehensive, re-annotation of the TB-Dense corpus, covering all possible event pairs. The dataset includes seven relation types: \textit{before}, \textit{after}, \textit{includes}, \textit{is-included}, \textit{equal}, \textit{overlap}, and \textit{vague}. Temporal relations are established based on event start times, end times, and durations. Notably, the \textit{vague} relation indicates that the temporal relation cannot be determined from the provided context or where annotators disagree, and it is crucial for \textit{complete} annotation, as it confirms that the pair was considered during annotation and deemed inconclusive, rather than ignored.

While NarrativeTime provides an exhaustively annotated dataset, it follows complex annotation guidelines similar to those of TB-Dense. MATRES refines these guidelines by focusing on a subset of events, annotating relations only based on event start times, and reducing the label set to \textit{before}, \textit{after}, \textit{equal}, and \textit{vague}. These refinements improve inter-annotator agreement and offer a more accessible setting for the task. However, MATRES is not exhaustively annotated. To bridge this gap, we develop \App{}, a dataset that adopts the refined MATRES scheme while ensuring complete coverage of \textit{all} event pairs across entire texts. Further details are provided in §\ref{section:dataset}.

\subsection{Zero-Shot Methods}
\label{section:background:models}

% The temporal relation extraction (TRE) task has traditionally relied on local pairwise methods, where the model processes one event pair at a time. This approach may stem from the lack of gold annotations (§\ref{section:background:datasets}) required for global modeling. 
% Furthermore, TRE has predominantly relied on supervised approaches \cite{huang-etal-2023-classification, tan-etal-2023-event, niu-etal-2024-contempo}, which require training datasets that are challenging to obtain or create \cite{naik-etal-2019-tddiscourse}, especially for new domains lacking annotated resources.

Recent advancements in LLMs offer an opportunity to leverage their vast knowledge for zero-shot approaches \cite{10.5555/3600270.3601883}, enabling solutions without training data \cite{zhao-etal-2023-pre}. However, few studies have explored LLMs for temporal relation extraction in zero-shot settings. The most notable and best-performing approach is by \citet{yuan-etal-2023-zero}, who applied a simple zero-shot chain-of-thought (CoT) method. In this method, the model is sequentially asked about each possible relation for a given event pair (e.g., ``Is event-a before event-b?''; if ``no,'' then ``Is event-a after event-b?'') until the model answers ``yes.'' We use \citet{yuan-etal-2023-zero} method as the zero-shot baseline in our experiments. Another effort by \citet{chan-etal-2024-exploring} experimented with prompt engineering and in-context learning. Both methods employed a pairwise approach and achieved suboptimal results on the MATRES and TB-Dense datasets. Additionally, the pairwise approach makes these methods cost- and time-inefficient. 

% Finally, determining the temporal relation between two events first requires understanding their temporal order. Several studies have demonstrated that planning before solving \cite{wang-etal-2023-plan} helps improve LLM performance in zero-shot across a range of tasks.

%Ultimately, deeming unsupervised methods as suboptimal for temporal relation extraction \cite{alsayyahi-batista-navarro-2023-timeline, wei-etal-2024-llms, niu-etal-2024-contempo, chan-etal-2024-exploring}.

The main goal in this work is to provide a more efficient and effective alternative to pairwise approaches by processing the entire document globally in a single step (see §\ref{section:model}).

\section{The \App{} Dataset}
\label{section:dataset}
\App{}\footnote{Released under a custom license that permits free academic use (see Appendix~\ref{append:dataset-license}).} is built following the MATRES \cite{ning-etal-2018-multi} approach (§\ref{section:background:datasets}); however, instead of annotating events only in consecutive sentences, the annotation is \textit{complete}, covering all event pairs across the entire document.
\App{} consists of a set of 30 human-generated English news summaries (\url{Newser.com}), derived from the Multi-News dataset \cite{fabbri-etal-2019-multi}. We select summaries that depict major events (e.g., a presidential visit abroad, a mass shooting, a major earthquake), as these are typically rich in informative sub-event mentions that describe the event timeline. Each summary contains a set of event mentions, with every pair assigned one of the following relations: \textit{before}, \textit{after}, \textit{equal}, or \textit{vague}.
We now describe \App{}'s annotation process (§\ref{section:background:annot-process}) along with dataset statistics (§\ref{section:dataset:statistics}).

\subsection{Annotation Process}
\label{section:background:annot-process}

For the annotation process, we hired three non-expert, native English-speaking annotators (students) to label 30 news summaries (\textasciitilde500 words each) for temporal relations (\textit{before}, \textit{after}, \textit{equal}, or \textit{vague}) between salient events, following MATRES guidelines and using the EventFull tool \cite{eirew2024eventfullcompleteconsistentevent}.\footnote{The complete annotation guidelines are available within the EventFull annotation tool and GitHub repository \url{https://github.com/AlonEirew/EventFull}.} Starting from \textasciitilde60 auto-detected event mentions per document, extracted using \citet{cattan-etal-2021-cross-document}, annotators selected 15–18 salient events for full-pair annotation, balancing coverage and annotation quality, as prior work shows agreement declines beyond this range for non-expert annotators \cite{eirew2024eventfullcompleteconsistentevent}. Final labels were determined by majority vote; in cases of disagreement, the label was set to vague. Further details on the annotation methodology and protocol are provided in Appendix~\ref{appx:annot-process}.

\begin{table}[!t]
    \centering
    % \small
    \resizebox{0.65\columnwidth}{!}{
    \begin{tabular}{@{}l|rr|r@{}}
        \toprule
        & \makecell{Train} & \makecell{Test} & \makecell{All} \\
        \midrule
        Documents & 20 & 10 & 30 \\
        Events & 319 & 151 & 470 \\
        \midrule
        \textit{before} & 1,119 & 419 & 1,538 \\
        \textit{after} & 916 & 431 & 1,347 \\
        \textit{equal} & 90 & 60 & 150 \\
        \textit{vague} & 276 & 172 & 448 \\
        \midrule
        Total Relations & 2,401 & 1,082 & 3,483 \\
        \bottomrule
    \end{tabular}}
    \caption{\App{} dataset statistics.}
    \label{tab:stats}
\end{table}

\subsection{Dataset Statistics and Comparison}
\label{section:dataset:statistics}

Table~\ref{tab:stats} summarizes the \App{} dataset's statistics. Overall, \App{} consists of 30 documents, corresponding to 470 event mentions and 3,483 relations. 
Table~\ref{tab:stats_all} in Appendix~\ref{append:additional-figures}, presents the statistics of prominent datasets for the temporal relation extraction task alongside \App{}.

The agreement among our annotators averaged 0.72 kappa \cite{kappa-1973}, corresponding to substantial agreement and is comparable to that of TB-Dense \cite{chambers-etal-2014-dense} (0.56$\kappa$–0.64$\kappa$), NarrativeTime \cite{rogers-etal-2024-narrativetime} (0.68$\kappa$), TDD-Manual \cite{naik-etal-2019-tddiscourse} (0.69$\kappa$), and MATRES \cite{ning-etal-2018-multi} (0.84$\kappa$). Additionally, to verify annotation accuracy, one of the authors re-annotated 50 random pairs, with 46 matching the majority vote of the annotators, further confirming the high quality of the annotations.

% Notably, both MATRES \cite{ning-etal-2018-multi} and \App{} define temporal relations based on event start times without considering duration, using four relation types (\textit{before}, \textit{after}, \textit{equal}, and \textit{vague}). In contrast, TimeBank-Dense \cite{chambers-etal-2014-dense} and NarrativeTime \cite{rogers-etal-2024-narrativetime} annotate six and seven relation types, respectively, incorporating \textit{includes} and \textit{is-included} and considering both start and end times as well as event duration. These distinctions create different challenges for models, with TimeBank-Dense and NarrativeTime presenting a more complex task. This difference is also reflected in the distinct relation distributions between the two groups, as shown in Table~\ref{tab:stats}.

% , and in contrast to all other relevant datasets (listed in Appendix~\ref{append:additional-figures}, Table~\ref{tab:stats_all})

Finally, we assess whether transitivity can compensate for the limited annotation scope in datasets like MATRES and TB-Dense, where only consecutive-sentence pairs are annotated. Using the NarrativeTime dataset, we consider only intra- and consecutive-sentence relations, then apply a transitive closure algorithm \cite{warsheall-1962} to infer additional links. While some long-distance relations are recovered, most inferred relations remain local and sparse (as illustrated in Figure~\ref{fig:nt_sentdiff} of Appendix~\ref{append:additional-figures}), further highlighting the importance of exhaustive annotation.

%as mentioned, our motivation for developing \App{} was to provide complete annotations within each document, similar to NarrativeTime (§\ref{section:background:datasets}). However, in datasets such as MATRES and TB-Dense, where annotation is complete only between consecutive sentences, event pairs may be inferred through transitivity rules. The extent to which this automatic inference scales, however, remains unclear. To investigate this, we analyze the NarrativeTime dataset by considering all relations within the same sentence or between consecutive sentences. We then apply a transitive algorithm to infer additional relations and assess how many can be recovered beyond a single sentence. Our analysis shows that while some long-distance relations are recovered, most inferred relations remain within close proximity and occur infrequently. This finding highlights the importance of exhaustive annotation. A full analysis is provided in Appendix~\ref{appx:trans-rels}.

% in constructing more complete and accurate story timelines.

\section{Zero-Shot Temporal Graph Generation}
\label{section:model}
% In this section, we outline our approach to zero-shot global temporal graph generation. 
%We begin with a vanilla global approach (ZSL-Global), refine it using a chain-of-thought planning approach (ZSL-Timeline), and then describe two methods for enforcing consistency — ZSL-SelfConsistency and ZSL-GraphConsistency — on the generated temporal graphs.

\subsection{Prompt Structure}

Our zero-shot approach, referred to as \textit{GlobalConsistency}, begins with a straightforward yet powerful idea: prompting an LLM to generate the full temporal graph of a document in a single call (Figure~\ref{fig:figure1}). The process starts with a general instruction outlining the task. We then employ a two-step procedure, motivated by the observation that directly prompting the model to classify all event pair relations results in a more inconsistent outputs. To address this, and inspired by reasoning-based prompting techniques \cite{wang-etal-2023-plan, sun-etal-2024-pearl}, we first prompt the model ([1] in Figure~\ref{fig:figure1}) to construct a free-form timeline that summarizes the temporal flow of the marked events. This primes the model with a broader understanding of event order before making explicit classification decisions. We then instruct the model to predict temporal relations between all event pairs. The input includes the full document with event mentions highlighted using angle brackets and unique identifiers (e.g., \texttt{<attack(7)>}), followed by a list of all possible event pairs. 

For the output, we instruct the model to represent relations as a graph, where events serve as nodes and relations as edges, formatted in the DOT language \cite{Gansner2006DrawingGW}, which helps suppress free-text explanations and facilitates parsing (an example of the generated timeline is presented in Appendix~\ref{append:additional-figures}, Figure~\ref{fig:timeline-output}).

In documents containing many events that may exceed the model’s input capacity, we generate the complete set of pairs and split them evenly for separate processing. Each split receives the same instructions and the full report with all event mentions marked in it, this is followed by only the relevant subset of event pairs for that split. The predictions are then merged back in post-processing (Further details are provided in Appendix~\ref{appx:experiment:details})

%If the list exceeds 200 pairs, it is split into batches and processed separately, then merged in post-processing (see Appendix~\ref{appx:experiment:details}).

\subsection{Post Process}
LLMs are inherently stochastic and may produce different labels for the same input when run multiple times, leading to unstable outputs, especially for ambiguous event pairs. To address this, inspired by self-consistency methods \cite{wang2023selfconsistency} and temporal graph consistency optimization techniques \cite{ning-etal-2018-joint}, we run the model \(M = 5\) times per document, as experimental results show that performance saturates after five generations (see Figure~\ref{fig:generation_increase} in Appendix~\ref{appx:experiment:details}), and aggregate the predicted relation labels into a distribution \( p_{ij} \in \mathbb{R}^{|\mathcal{R}|} \) for each event pair \( (e_i, e_j) \in \mathcal{E} \times \mathcal{E} \), where \( \mathcal{E} \) denotes the set of events, and \( p_{ij}^r \) represents the empirical likelihood of label \( r \in \mathcal{R} \) across runs. ([2,3] in Figure~\ref{fig:figure1}).

We then apply the Integer Linear Programming (ILP) formulation of \citet{ning-etal-2018-joint} to obtain a globally consistent graph. Specifically, we define binary variables \( \mathcal{I}_r(i, j) \in \{0,1\} \) for each relation \( r \) and pair \( (e_i, e_j) \), and optimize for enforcing key structural constraints: uniqueness (only one relation per pair), symmetry (e.g., if \( r = \texttt{BEFORE} \), then its inverse holds for the reverse pair), and transitivity (e.g., if \( A \rightarrow B \rightarrow C \), then \( A \rightarrow C \)) ([4] in Figure~\ref{fig:figure1}). The result is the optimal temporal graph that maximizes model confidence while ensuring global coherence. Further details are provided in Appendix~\ref{appx:formal-zsl-global}.

% Finally, to assess the contribution of each component, we define three ablation variants (§5.3):
% \textbf{ZSL-Global}, which skips timeline generation and directly predicts the graph;
% \textbf{ZSL-Timeline}, which omits global consistency optimization; and
% \textbf{ZSL-SelfConsistency}, which replaces optimization with majority voting.
% These variants isolate the effect of timeline-based reasoning and consistency enforcement.

\section{Experimental Setting}
\label{section:experiment}

We describe the datasets and models used in our experiments. Technical details are in Appendix~\ref{appx:experiment:details}.
% In this section, we detail our experimental setup, as presented in Table~\ref{tab:models-perform}, including the dataset used for evaluation (§\ref{section:experiment:datasets}), and baseline models (§\ref{section:experiment:baselines}). 

\begin{table*}[!t]
    \centering
    \resizebox{0.7\textwidth}{!}{
    \begin{tabular}{@{}l|cc|cc||cc|cc@{}}
        \toprule
        \textbf{} 
        & \multicolumn{4}{c||}{\textit{Non-Exhaustive}} 
        & \multicolumn{4}{c}{\textit{Exhaustive}} \\
        \cmidrule(lr){2-5} \cmidrule(lr){6-9}
         & \multicolumn{2}{c|}{\textbf{MATRES}} 
         & \multicolumn{2}{c||}{\textbf{TB-Dense}} 
         & \multicolumn{2}{c|}{\textbf{NT-6}} 
         & \multicolumn{2}{c}{\textbf{\App{}}} \\
         & F1 & TI & F1 & TI & F1 & TI & F1 & TI \\
        \midrule
        \multicolumn{9}{c}{\textbf{\textit{Supervised} SOTA Pairwise Models}} \\
        \midrule
        \textasteriskcentered~RoBERTa (\citeauthor{tan-etal-2023-event}) & 80.4 & 24 & 60.5 & 107 & 59.3 & 105 & 73.6 & 143 \\
        \textasteriskcentered~Bayesian (\citeauthor{tan-etal-2023-event}) & \textbf{82.7} & 16 & \textbf{65.0} & 87 & 64.9 & 203 & 78.7 & 166 \\
        + Constraints & -- & -- & -- & -- & \textbf{65.6} & 0 & \textbf{80.7} & 0 \\
        \midrule
        \multicolumn{9}{c}{\textbf{\textit{Zero-Shot} Prompting with \textit{GPT-4o}}} \\
        \midrule
        CoT (\citeauthor{yuan-etal-2023-zero}) & 56.6 & -- & \textbf{42.8} & -- & 49.3 & 461 & 67.2 & 374 \\
        \midrule
        ZSL-Global (Ours) & 59.0 & 73 & 37.7 & 250 & 48.4 & 300 & 62.3 & 161 \\
        ZSL-Timeline (Ours) & 58.4 & 81 & 39.1 & 225 & 52.2 & 309 & 68.5 & 157 \\
        \midrule
        SelfConsistency (Ours) & 60.1 & 50 & 41.2 & 122 & 55.6 & 305 & 71.0 & 128 \\
        GlobalConsistency (Ours) & \textbf{63.0} & 0 & \textbf{42.8} & 0 & \textbf{58.4} & 0 & \textbf{73.6} & 0 \\
        \midrule
        \multicolumn{9}{c}{\textbf{\textit{Zero-Shot} Prompting with \textit{DeepSeek-R1}}} \\
        \midrule
        CoT (\citeauthor{yuan-etal-2023-zero}) & \textbf{70.3} & -- & \textbf{50.8} & -- & 57.9 & 360 & 78.4 & 254 \\
        \midrule
        ZSL-Global (Ours) & 61.0 & 82 & 44.6 & 276 & 57.0 & 262 & 70.5 & 167 \\
        ZSL-Timeline (Ours) & 59.0 & 82 & 44.4 & 261 & 59.4 & 185 & 74.6 & 90 \\
        \midrule
        SelfConsistency (Ours) & 61.2 & 58 & 46.8 & 152 & 62.1 & 144 & 78.7 & 79 \\
        GlobalConsistency (Ours) & 66.4 & 0 & 49.0 & 0 & \textbf{64.1} & 0 & \textbf{79.2} & 0 \\
        \bottomrule
    \end{tabular}}
    \caption{F1 and Transitive Inconsistency (TI) scores of all models on four datasets, grouped into \textit{Non-Exhaustive} Annotation (MATRES, TB-Dense) and \textit{Exhaustive} Annotation (NT-6, \App{}). We use the F1 definition from \citet{ning-etal-2019-improved}, and compute the average number of TI edges per test document by applying a transitive closure algorithm \cite{warsheall-1962} and counting transitive contradictions. (\textasteriskcentered) For MATRES and TB-Dense with supervised models, we report results from \citet{tan-etal-2023-event} as our reproductions were slightly lower; TI is based on our retrained models. Constraints are not reported for these models as they did not improve results. (--) TI is not reported for CoT on MATRES and TB-Dense, as it only predicts gold-labeled pairs and cannot construct a complete graph. Computing TI for CoT would require multiple generations over the full set of pairs, which is prohibitively expensive (see Table~\ref{tab:costs}). Further details are provided in Appendix~\ref{appx:main-res-details}.}
    \label{tab:models-perform}
\end{table*}

\subsection{Datasets}
\label{section:experiment:datasets}

In our experiments, we use our own \App{} and three additional datasets: MATRES, TB-Dense, and NarrativeTime.\footnote{Dataset license details are in Appendix~\ref{append:dataset-license}.}
Notably, TCR \cite{ning-etal-2018-joint} and TDD-Manual \cite{naik-etal-2019-tddiscourse}, two additional datasets for the TRE task, are excluded from our experiments as they omit the \textit{vague} relation. Since we generate relations for all possible event pairs, the \textit{vague} label is essential to avoid forcing incorrect relations when context is insufficient. 
Below, we provide details on the datasets used in our experiments. 
For our own \App{}, we use the first 10 documents as the test set and the remaining documents as the training set, while for all other datasets, we follow their predefined splits.

\textbf{MATRES.} In MATRES, only events within consecutive sentences are annotated. The dataset includes four relation types: \textit{before}, \textit{after}, \textit{equal}, and \textit{vague}, with temporal relations determined based on event start times. 
% We conduct experiments on MATRES twice: once using the full test set, referred to as \textit{all}, and once using a subset of the test documents containing 20 or fewer event mentions, referred to as \textit{small}. This subset allows us to assess model performance on documents with fewer events, based on our observation that performance varies with the number of event mentions in a document, as further discussed in §\ref{section:results}.

\textbf{TB-Dense.} Similar to MATRES, only events within consecutive sentences are annotated in the TB-Dense dataset. It includes six relation types, the four from MATRES plus \textit{includes} and \textit{is-included}. Temporal relations are determined based on event start and end times as well as their duration. 
% Like before, to assess model performance on documents with fewer events, we experiment with TB-Dense twice, once using the full test set, referred to as \textit{TB-Dense (all)}, and once with a subset of the test set containing only documents with 30 or fewer event mentions, referred to as \textit{TB-Dense (mid)}.

\textbf{NT-6.} The NarrativeTime (NT) dataset, previously introduced in §\ref{section:background:datasets}, features seven relation types, including the six from TB-Dense and the \textit{overlap} relation. 
However, we exclude the \textit{overlap} relation as it is incompatible for the temporal consistency methods, given that the symmetric counterpart was not annotated.
Additionally, NT documents contain an average of 50 events, corresponding to 1,200 relations, per document. Due to LLM context limits, we randomly select 18 events per NT document.

\subsection{Baseline and State-of-the-Art Models}
\label{section:experiment:baselines}

We compare our GlobalConsistency method with four models, reproducing state-of-the-art (SOTA) supervised models and a zero-shot chain-of-thought (CoT) baseline method.

\textbf{Bayesian \cite{tan-etal-2023-event}.} 
Bayesian-Translation is the current publicly available state-of-the-art pairwise model for temporal relation extraction. It leverages a COMET-BART encoder \cite{Hwang2020COMETATOMIC2O} and a graph translation model \cite{Balazevic2019MultirelationalPG} to incorporate prior knowledge from the ATOMIC commonsense knowledge base, refining event representations for relational embedding learning. Additionally, it employs a Bayesian framework to estimate the uncertainty of the learned relations.

% uses the RoBERTa-large \cite{zhuang-etal-2021-robustly} encoder to represent event pairs and a graph translation model \cite{Balazevic2019MultirelationalPG} to learn relational embeddings. The model applies a Bayesian framework to estimate the uncertainty of the learned relations.

%\textbf{Bayesian \cite{tan-etal-2023-event}.} A strong pairwise model for temporal relation extraction, similar in architecture to the above RoBERTa baseline. However, it replaces the RoBERTa encoder with a COMET-BART encoder \cite{Hwang2020COMETATOMIC2O}, and integrate prior knowledge from the ATOMIC commonsense knowledge base.

% \textbf{RoBERTa \cite{tan-etal-2023-event}.} A pairwise model that uses the RoBERTa-large \cite{zhuang-etal-2021-robustly} encoder to represent event pairs and a graph translation model \cite{Balazevic2019MultirelationalPG} to learn relational embeddings. The model applies a Bayesian framework to estimate the uncertainty of the learned relations.

\textbf{RoBERTa \cite{tan-etal-2023-event}.} A strong pairwise model for temporal relation extraction, similar in architecture to the Bayesian model described above, but replacing the COMET-BART encoder with a RoBERTa-large encoder \cite{zhuang-etal-2021-robustly}. 
%Unlike the Bayesian model, it learns relational embeddings without relying on prior knowledge from external sources. 
We use this model as it represents a strong, purely supervised approach, allowing for a direct comparison without the influence of external knowledge.

\textbf{Bayesian + Constraints.}  
We extend the Bayesian model with the temporal constraints optimization algorithm \cite{ning-etal-2018-joint}, the same algorithm used in our GlobalConsistency method, applying it at inference time to enable a more direct comparison with our methods.

% \textbf{Bayesian + Constraints.} 
%An extended version of the Bayesian model that applies transitive constraints optimization at inference time, following \cite{ning-etal-2018-joint}.
% An extension of the Bayesian-Trans model, where transitive constraints optimization is applied at inference, following \cite{ning-etal-2018-joint}.

\textbf{CoT \cite{yuan-etal-2023-zero}.} 
As a baseline model, we re-implemented the CoT model \cite{yuan-etal-2023-zero} using GPT-4o and DeepSeek-R1, replacing the original implementation, which used ChatGPT. To the best of our knowledge, this is the strongest zero-shot approach for temporal relation extraction. Unlike our method, which generates relations for all event pairs, the CoT baseline is applied only to event pairs with gold annotations, due to its high computational cost.

For evaluation, we report the F1 score on all datasets following the definition in \cite{ning-etal-2019-improved}, where the \textit{vague} relation is excluded from true positive predictions. Additionally, we report a Temporal Inconsistency (TI) measure by applying a transitive closure algorithm \cite{warsheall-1962} and counting transitive contradictions (further details in Appendix~\ref{appx:trans-inconsist}).
%—i.e., relations that violate the transitivity constraints defined by \citet{ning-etal-2018-joint}.

\subsection{Ablation Study Design}
To investigate the contribution of each component in our method to the overall performance, we design the following three ablation models:

\textbf{ZSL-Global.} This zero-shot learning (ZSL) configuration prompts the LLM once to generate the entire temporal graph directly, ommiting the instruction to generate the timeline of events before classification.

\textbf{ZSL-Timeline.} This ZSL configuration includes only the prompting step (with timeline generation) but omits the post-processing step that enforces global consistency.

\textbf{SelfConsistency.} This configuration replaces our global consistency optimization with a simpler self-consistency approach \cite{wang2023selfconsistency}, where the final label for each event pair is selected by majority vote from the five generated outputs.

\section{Results}
\label{section:results}

Our results are presented in Table~\ref{tab:models-perform}, with \textit{supervised} SOTA pairwise models shown in the upper section, and the results of our \textit{zero-shot} methods, using GPT-4o and DeepSeek-R1, shown in the lower section.\footnote{Details on additional models and datasets evaluated are provided in Appendix~\ref{appx:additional-tests}.} Overall, using GPT-4o, our GlobalConsistency approach (§\ref{section:model}) outperforms the CoT baseline \cite{yuan-etal-2023-zero} by a large margin across all datasets except TB-Dense.
Using DeepSeek-R1, our GlobalConsistency method outperforms the CoT baseline on the densely annotated datasets, NT-6 and \App{}, but shows lower performance on the sparsely annotated datasets, MATRES and TB-Dense (further analyzed in §\ref{section:results:quality}). However, the improved performance of the CoT method comes at a significant cost, \textasciitilde7$\times$ more expensive (Table~\ref{tab:costs}), and requires more time, particularly in the DeepSeek-R1 experiments (Table~\ref{tab:times}).
Furthermore, time and cost differences between CoT and our method do not scale linearly with graph size. This is evident in Table~\ref{tab:costs}, where NT-6, with only two more events per document than \App{}, incurs higher costs with CoT.

\begin{figure}[t!]
    \centering
    \includegraphics[width=0.95\linewidth]{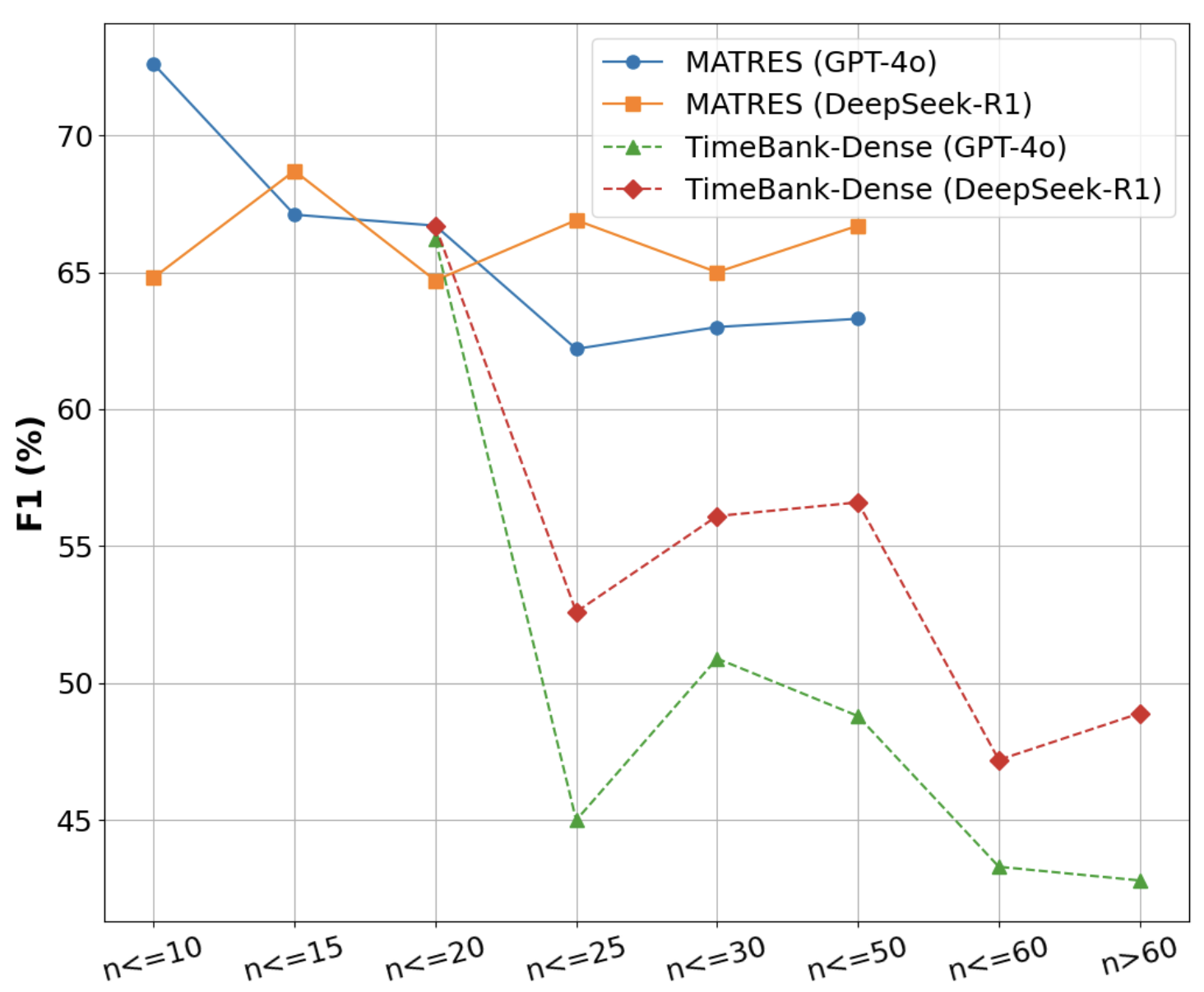}
    \caption{Impact of event count per document on GlobalConsistency performance, evaluated on MATRES and TB-Dense. The x-axis is cumulative, and the y-axis shows the F1 score per subset.}
    \label{fig:event-buckets}
\end{figure}

Notably, on the dense datasets, NT-6 and \App{}, GlobalConsistency using DeepSeek-R1 matches supervised models (79.2 vs. 80.7 for OmniTemp, 64.1 vs. 65.6 for NT-6), while producing more consistent graphs with lower transitive inconsistency (TI) scores reported in the table. Moreover, our approach requires no training data and does not rely on a substantial external commonsense knowledge base (as required by the Bayesian-Trans model for example), which may not be applicable across many domains and languages. This positions GlobalConsistency as an appealing zero-shot alternative for TRE in scenarios where labeled training data or comprehensive knowledge resources are rare or unavailable.

% We proceed with providing deeper observations through qualitative analysis.

\section{Discussion}
\label{section:results:quality}

\paragraph{Event Mentions Count.}
We investigate how the number of events in a document impacts the performance of our GlobalConsistency method. Our hypothesis is that models encoding global information are more sensitive to event count, as they must process more information simultaneously. In contrast, pairwise methods, which consider one event pair at a time, are likely less affected.  
Figure~\ref{fig:event-buckets} shows MATRES and TB-Dense documents grouped by increasing event counts.\footnote{The other datasets we experimented with contain a limited number of events per instance.}  
With the exception of DeepSeek-R1 on MATRES, which demonstrates resilience to large event counts, the results show a performance decline as the number of events increases. This supports our hypothesis and may help explain the performance gap between the CoT method and the ZSL-Global variant in most tests.\footnote{In TB-Dense, performance drops sharply for documents with over 25 events. For further analysis, see Appendix~\ref{appx:fig-expl:fig2}.}

\begin{table}[!t]
    \centering
    \resizebox{\columnwidth}{!}{
    \begin{tabular}{@{}l|c|c|c|c@{}}
        \toprule
        & \multicolumn{2}{c|}{\textbf{CoT}}  & \multicolumn{2}{c}{\textbf{GlobalConsistency}} \\
        \toprule
        & GPT-4o & DeepSeek-R1 & GPT-4o & DeepSeek-R1 \\
        \midrule
        MATRES & 50 & 69 & 6 & 9\\
        TB-Dense & 71 & 99 & 9 & 17 \\
        NT-6 & 15 & 21 & 2 & 3 \\
        \App{} & 12 & 16 & 2 & 3 \\
        \bottomrule
    \end{tabular}}
    \caption{Approximate costs (USD) for the full dataset are shown. For GlobalConsistency, the cost reflects five generations of the complete set of relations. For the CoT method, to reflect a real-world scenario, we generate the complete set of relations once—rather than just the gold ones. Costs are computed using token counts (via OpenAI’s \texttt{tiktoken}, and DeepSeek official tokenizer) and official model pricing.}
    \label{tab:costs}
\end{table}

\begin{table}[!t]
    \centering
    \resizebox{\columnwidth}{!}{
    \begin{tabular}{@{}l|c|c|c|c@{}}
        \toprule
        & \multicolumn{2}{c|}{\textbf{CoT}}  & \multicolumn{2}{c}{\textbf{GlobalConsistency}} \\
        \toprule
        & GPT-4o & DeepSeek-R1 & GPT-4o & DeepSeek-R1 \\
        \midrule
        MATRES & 297 & 2,789 & 250 & 363\\
        TB-Dense & 426 & 4,004 & 360 & 521 \\
        NT-6 & 89 & 838 & 75 & 110 \\
        \App{} & 70 & 658 & 60 & 86 \\
        \bottomrule
    \end{tabular}}
    \caption{Time (in minutes) to generate the full set of temporal relations for each test set. For GlobalConsistency, this includes five generations; for CoT, it reflects a single pass over all relations (not just gold) to simulate real-world use.}
    \label{tab:times}
\end{table}

\begin{figure}[t]
    \centering

    % First image with caption ONLY ABOVE
    \parbox{\linewidth}{ % Wrap in parbox so caption is above
        \centering{\footnotesize (a) \App{}.} \\ % Manually add caption above
        \includegraphics[width=0.98\linewidth]{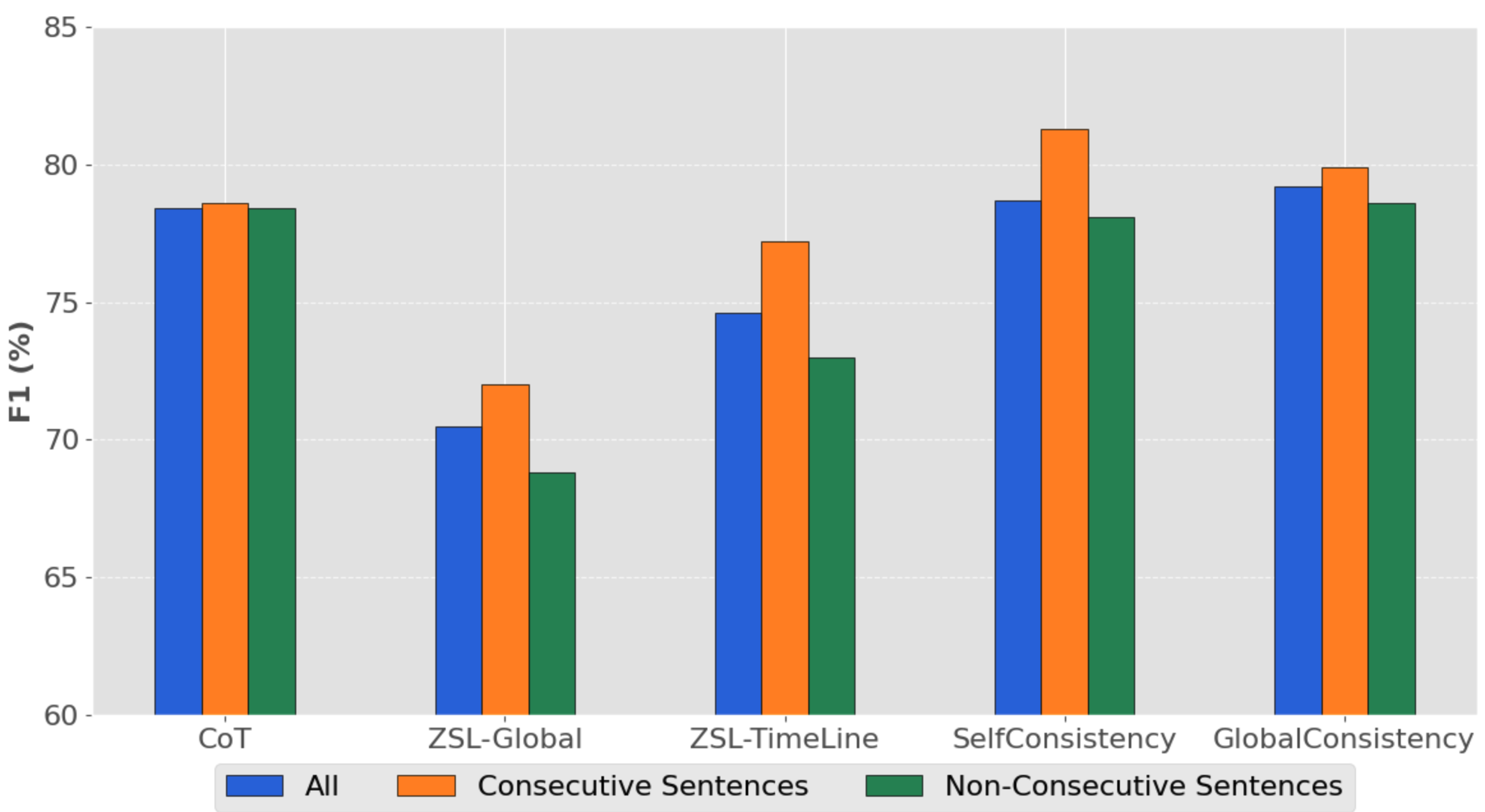}
    }

    \vspace{0.5cm} % Adjust spacing between figures

    % Second image with caption ONLY ABOVE
    \parbox{\linewidth}{
        \centering{\footnotesize (b) NarrativeTime.} \\ % Manually add caption above
        \includegraphics[width=0.98\linewidth]{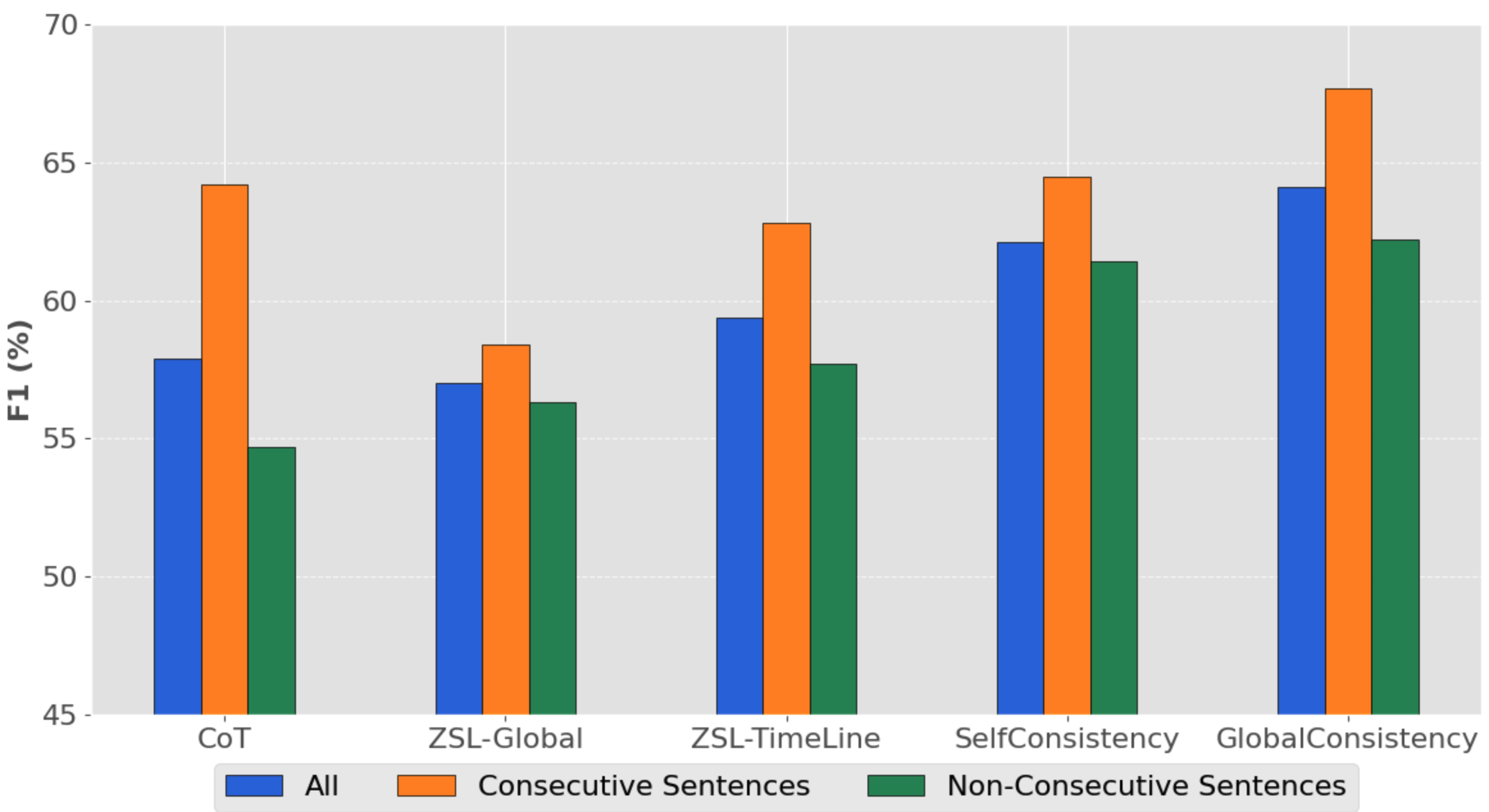}
    }

    % Main figure caption at the bottom
    \caption{DeepSeek-R1 model performance across different relation subsets: (1) consecutive-sentence event pairs, (2) non-consecutive-sentence event pairs, and (3) full-document event relations. A similar figure for GPT-4o is presented in Figure~\ref{fig:cs-sent-expr-gpt} in Appendix~\ref{append:additional-figures}.}
    \label{fig:cs-sent-expr-ds}
\end{figure}

\paragraph{Event Pair Distance.}
% From Table \ref{tab:models-perform} we learn that our initial version ZSL-Global outperforms the CoT baseline on MATRES but underperforms on \App{}. 
We examine whether the annotation distance restriction, where events are annotated only if they are at most one sentence apart, as in MATRES and TB-Dense, can affect model evaluation. To explore this, we evaluate all zero-shot methods on three subsets of \App{} and NT-6: the full dataset, event pairs with a sentence distance of at most one (consecutive sentences like in MATRES and TB-Dense), and event pairs with a sentence distance greater than one (non-consecutive sentences). See Figure~\ref{fig:cs-sent-expr-ds}.
%\footnote{For a fair comparison, we compare CoT to our methods before applying global consistency, isolating its performance from that achieved through transitive constraints, whose effectiveness depends on the quality of the input relations.}

%ZSL-Global and ZSL-Timeline use a single-turn prompting approach, while ZSL-SelfConsistency and ZSL-GlobalConsistency extend this by interacting with the LLM multiple times. To ensure a fair comparison with the baseline CoT method, which also relies on a single LLM call, we include only the single-call methods in this evaluation.

Our findings show that on the four-relation \App{} dataset, the CoT baseline performs consistently across all sentence distances, while our global methods achieve higher performance on consecutive-sentence pairs. In contrast, on the more challenging six-relation NT-6 dataset, CoT performs notably better on consecutive-sentence pairs than on long-distance pairs. These findings highlights the importance of document-level annotations for reliable evaluation of temporal relation classification—especially in zero-shot settings, where models cannot realistically rely on distribution patterns in the annotations.

%Moreover, they may explain why ZSL-Global slightly outperforms ZSL-Timeline on the consecutive-sentence datasets (MATRES and TB-Dense), despite lacking intermediate timeline reasoning. This suggests that the performance gap may reflect dataset bias, and highlights the need for evaluating models on fully annotated, long-range temporal graphs.

%Our findings show that on the four-relation \App{} dataset, the CoT baseline performs consistently across all sentence distances, while our global methods perform significantly better on consecutive-sentence relations. This may explain why ZSL-Global outperforms the ZSL-Timeline approach on MATRES. Interestingly, the opposite trend is observed in the more challenging six-relation NT-6 dataset, where the CoT baseline performs much better on consecutive-sentence relations compared to its performance on the non-consecutive-sentence subset. These findings highlights the importance of document-level annotations for reliable evaluation of temporal relation classification—especially in zero-shot settings.
%, where models cannot realistically rely on distribution patterns in the annotations.

\begin{figure}[t]
    \centering

    % First image with caption ONLY ABOVE
    \parbox{0.95\linewidth}{ % Wrap in parbox so caption is above
        \centering{\footnotesize (a) NarrativeTime Vs. TimeBank-Dense.} \\ % Manually add caption above
        \includegraphics[width=0.98\linewidth]{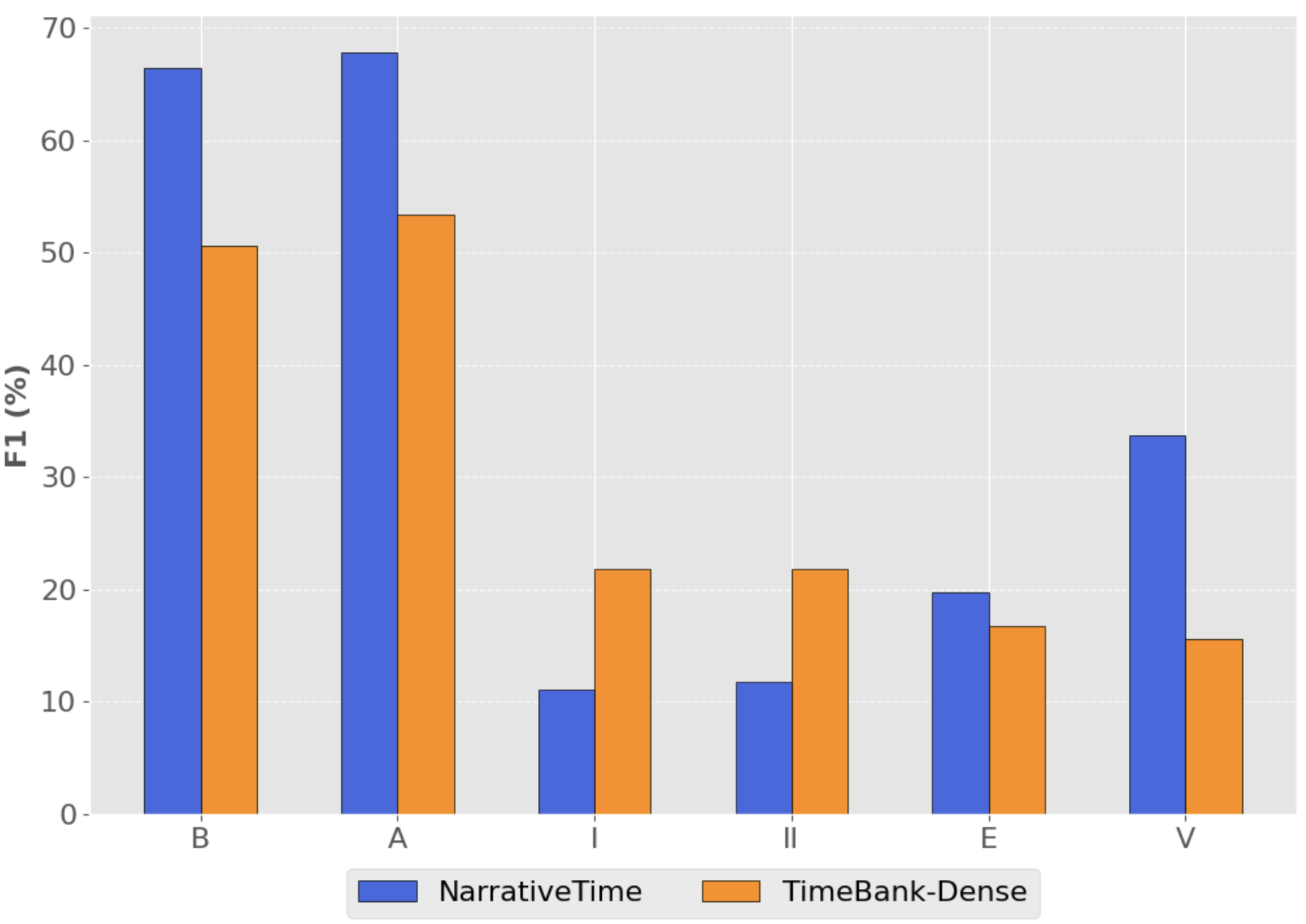}
    }

    \vspace{0.5cm} % Adjust spacing between figures

    % Second image with caption ONLY ABOVE
    \parbox{0.95\linewidth}{
        \centering{\footnotesize (b) \App{} Vs. MATRES.} \\ % Manually add caption above
        \includegraphics[width=0.98\linewidth]{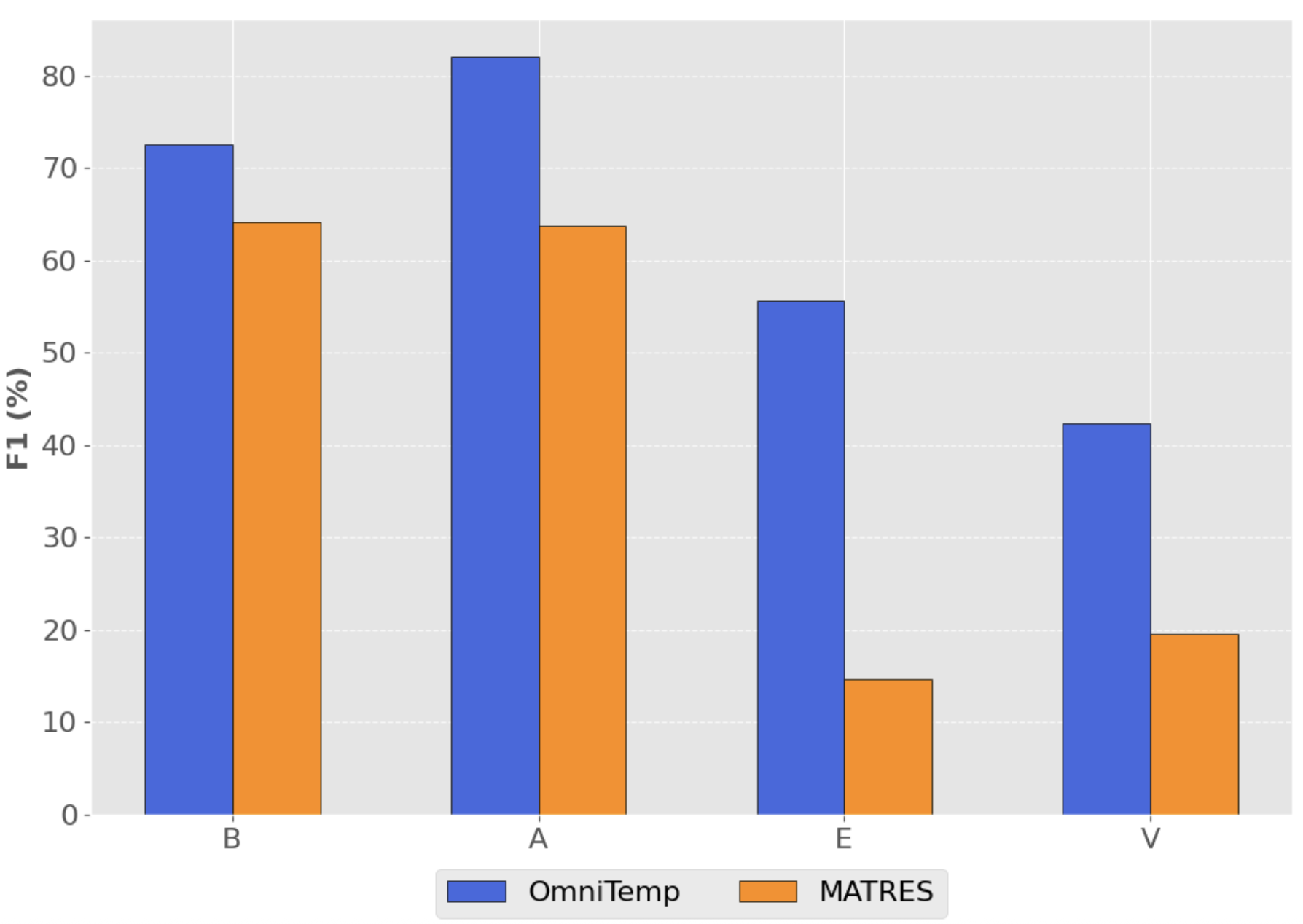}
    }

    % Main figure caption at the bottom
    \caption{We examine the performance of our prompting method (i.e., ZSL-Timeline) by relation type across two groups of datasets with similar annotation schemes: six-label datasets (TB-Dense and NT-6) and four-label datasets (MATRES and \App{}), using DeepSeek-R1. Similar results are observed with GPT-4o, as presented in Figure~\ref{fig:per-rel-expr-gpt} in Appendix~\ref{append:additional-figures}. The relation labels are: A = \textit{after}, B = \textit{before}, I = \textit{includes}, II = \textit{is-included}, E = \textit{equal}, and V = \textit{vague}.}
    \label{fig:per-rel-expr-ds}
\end{figure}

\paragraph{Label Inconsistency.}
The performance gap between our methods and the supervised models varies across datasets, being more pronounced in MATRES and TB-Dense than in NT-6 and OmniTemp. To better understand this gap, we analyze the ZSL-Timeline variant (chosen to isolate the model's performance without the influence of post-processing) by examining results per label and grouping datasets with similar label sets, as shown in Figure~\ref{fig:per-rel-expr-ds}. Our ZSL-Timeline method performs significantly worse on MATRES and TB-Dense than on OmniTemp and NT-6.
%The four datasets are grouped into two charts: one for six-label datasets (TB-Dense and NT-6) and one for four-label datasets (MATRES and OmniTemp). Our ZSL-Timeline method performs significantly worse on MATRES and TB-Dense than on OmniTemp and NT-6.

To investigate this further, we examine label consistency in documents and event pairs shared between TB-Dense and MATRES, which annotated the same corpus. There are 983 such event pairs. While these datasets follow different annotation guidelines, certain labels should remain consistent. For instance, if an event pair is labeled \textit{equal} in TB-Dense, indicating that both the start and end times of the two events are the same, then the relation should also be \textit{equal} in MATRES. Measuring consistency across the four shared relations, we find strong agreement for \textit{before} and \textit{after}, with \textit{before} being the most consistently annotated. However, significant inconsistencies were evident in \textit{vague} and \textit{equal} (detailed results are provided in Appendix~\ref{appx:consist-eval}).  
Since in zero-shot settings the model is not trained on a dataset, it does not learn dataset-specific annotation biases. The annotation inconsistency between MATRES and TB-Dense may partly explain the performance drop on these datasets, particularly for \textit{vague} and \textit{equal} relations, as well as the lower performance on \textit{after} compared to \textit{before}.

This analysis, together with the pair distance analysis, may help explain the gap observed between the zero-shot and supervised methods on MATRES and TB-Dense, raising a broader question about the reliability of evaluating zero-shot approaches on partially annotated or inconsistent resources.

\section{Conclusion}

In this work, we introduced a novel zero-shot LLM approach for temporal relation extraction that generates the entire temporal graph at once. 
Our method moves beyond traditional pairwise approaches, which suffer from computational inefficiency and lack of global consistency. To ensure temporal consistency, we incorporated temporal constraints optimization, significantly improving both accuracy and efficiency while generating relations completely free of inconsistencies. Our results show that zero-shot LLMs, when prompted to generate the timeline of events in free-form language before assigning labels to event pairs and extended with a global constraints algorithm, can serve as a competitive alternative to supervised models, especially in low-resource or cross-domain settings where training data is scarce. Additionally, we introduced \textit{\App{}}, a new dataset with complete annotations for all event pairs, following the refined annotation guidelines of MATRES. By providing gold labels for every event pair in a document, this dataset enables a fair evaluation of zero-shot approaches.

\section*{Limitations}

While our proposed zero-shot temporal graph generation approach demonstrates significant advantages over pairwise methods, several limitations remain that warrant further investigation.

First, closed LLMs such as GPT-4o and DeepSeek-R1 do not disclose their training data. Therefore, results on the three datasets we investigate may be affected by potential data contamination if their test sets were included in the training phase. However, \App{} is a completely new resource that is not yet publicly available, ensuring uncontaminated results.

Second, although self-consistency prompting mitigates stochasticity to some extent, the model’s responses can still be inconsistent, especially when handling long-distance temporal dependencies or ambiguous event relations.

Finally, the computational cost of using LLMs for large-scale inference remains a challenge. While our approach significantly reduces costs compared to pairwise methods, generating a full temporal graph for documents with many events can still be time-intensive and expensive.

% Fourth, in this research, we present our results on GPT-4o; however, we expect similar conclusions with other equivalent LLMs.

% our dataset, \textbf{OmniTemp}, provides exhaustive event-event relation annotations following the MATRES-style four-relation schema (\textit{before, after, equal, vague}), considering only the start time of events. As a result, it may not represent all TRE tasks, such as those requiring the \textit{includes} relation or those that also consider event end times and durations.

Despite these limitations, our study highlights promising directions for leveraging LLMs in structured event reasoning and lays the groundwork for future improvements in temporal relation extraction.

\section*{Acknowledgments}
This work was supported by the Israel Science Foundation (grant no. 2827/21), and by funding from the Israeli Planning and Budgeting Committee (PBC).

We also used AI-based assistance (ChatGPT) for grammar and spelling corrections during manuscript preparation.

% Bibliography entries for the entire Anthology, followed by custom entries
%\bibliography{anthology,custom}
% Custom bibliography entries only
\bibliography{custom}

\appendix

\section{Experimental Details}
\label{appx:experiment:details}

\begin{figure}[t!]
\centering
\includegraphics[width=0.9\columnwidth]{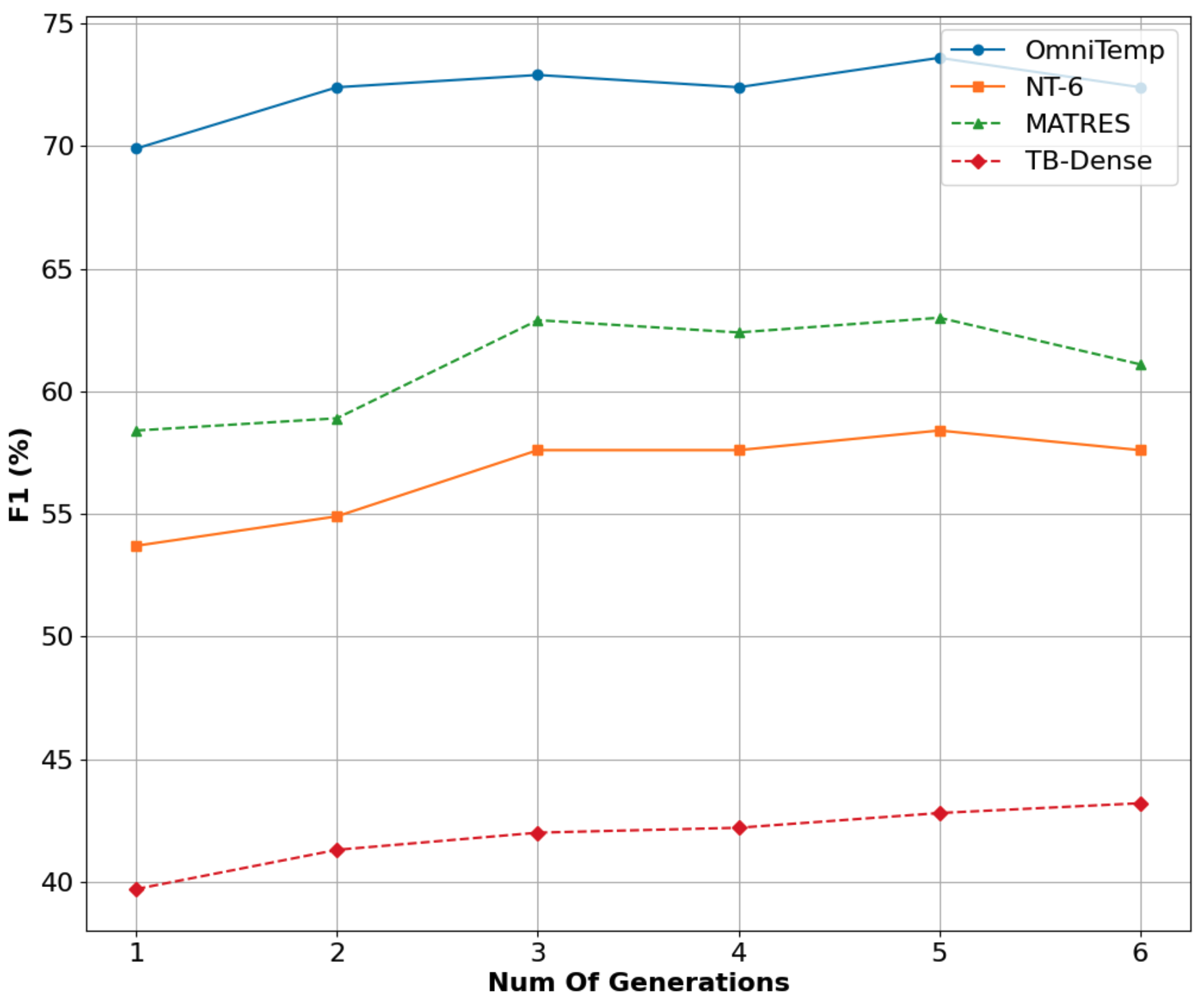}
\caption{Effect of increasing the number of generated instances and applying GlobalConsistency with GPT-4o. Results show improved performance, with saturation observed after about five generations in most datasets.}
\label{fig:generation_increase}
\end{figure}

For all supervised model experiments, we follow the experimental setup of \citet{tan-etal-2023-event}. To this end, we conducted a grid search to determine the optimal hyperparameters and embedding dimensionality for each test. Each training episode was run for 50 epochs on a single A100 GPU,\footnote{Experiment GPU time varies depending on the size of the training set, ranging from 1 to 20 hours for a full training episode.} with the best-performing epoch on the development set selected for evaluation. For the GPT-4o experiments, we use `gpt-4o-2024-08-06' version through OpenAI API, and used Together.ai API, for the DeepSeek-R1 experiments. We set the number of generations to five, based on tuning experiments with GPT-4o on the \App{} and NT-6 development sets (illustrated in Figure~\ref{fig:generation_increase}). In all experiments, we provide the model with all event pairs combinations, and evaluate on the available gold labels. For the MATRES and TimeBank-Dense (TB-Dense) datasets, we evenly divide the set of pairs in documents containing more than 20 events. In TB-Dense, for documents exceeding 40 events, we further group the pairs into sets of 100.
Finally, In cases the generation missed pairs or is malformed, we regenerate the document or its respective split. For temporal constraint optimization, we employ the Gurobi Optimizer \cite{gurobi}. Finally, the total experimental cost of this research—including CoT, ablation, and final results—using LLMs via OpenAI, Google, and Together.ai was approximately \$400 (USD).

\begin{table}[!t]
    \centering
    % \small
    \resizebox{0.75\columnwidth}{!}{
    \begin{tabular}{@{}l|cc@{}}
        \toprule
        \makecell{Model} & \makecell{NT-6} & \makecell{\App{}} \\
        \midrule
        DeepSeek-R1 & 64.1 & 79.2 \\
        DeepSeek-V3 & 55.1 & 74.4 \\
        GPT-o3-mini & 55.5 & 78.1 \\
        GPT-4o & 58.4 & 73.6 \\
        Llama-3.1 405B Instruct & 42.6 & 57.9 \\
        Llama 3.3 70B-Instruct & 40.7 & 44.3 \\
        Gemini-flash 2.0 & 32.1 & 47.3 \\
        \bottomrule
    \end{tabular}}
    \caption{Additional results of our GlobalConsistency approach when applying different LLMs, evaluated on the \App{} and NT-6 datasets.}
    \label{tab:add-models}
\end{table}

\begin{table}[!t]
    \centering
    \small
    % \resizebox{0.35\textwidth}{!}{
    \begin{tabular}{@{}l|c|c@{}}
        \toprule
        & \makecell{GPT-4o} & \makecell{DeepSeek-R1} \\
        \midrule
        GlobalConsistency & 71.2 & 64.9 \\
        \bottomrule
    \end{tabular}
    \caption{Results on the validation set of MAVEN-ERE \cite{wang-etal-2022-maven} on the sub-set of relations we selected.}
    \label{tab:maven_res}
\end{table}

\section{Additional Experiments}
\label{appx:additional-tests}
\subsection{Additional Tested LLMs}
\label{appx:addtional-tested-models}
Beyond our main experiments with GPT-4o and DeepSeek-R1, we also evaluated our model with additional LLMs, summarized in Table~\ref{tab:add-models} together with GPT-4o and DeepSeek-R1 for ease of comparison. GPT-o3-mini and DeepSeek-V3 achieved promising results on both NT-6 (55.1 and 55.5) and \App{} (74.4 and 78.1). Additionally, we tested our method with several contemporary models accessed via the Together.ai and Google Gemini APIs, including LLaMA (v3.1 405B, v3.3 70B) and Gemini (Flash-2.0, Pro-1.5). Our findings suggest that while all of these models (except Gemini Pro-1.5, which truncated the generation with the message: ``rest of the pairs are similar, and the logic should follow the timeline explanation.'') are capable of generating a complete set of temporal relations in a single step, they achieved much lower results. This indicates that our method for generating complete temporal graphs currently performs best with more advanced models.

\subsection{Additional Tested Dataset}
\label{appx:addtional-tested-datasets}
We conducted an additional experiment with our GlobalConsistency method on the MAVEN-ERE \cite{wang-etal-2022-maven} dataset, which introduces an additional domain, as it was curated from Wikipedia (in contrast to the newswire datasets used in our main experiments). MAVEN-ERE includes multiple relation categories, such as temporal, coreference, causal, and sub-events. In our setting, we used only the temporal relations portion of the dataset. For temporal relations, MAVEN-ERE defines six relation types: BEFORE, CONTAINS, OVERLAP, BEGINS-ON, ENDS-ON, and SIMULTANEOUS. To manage costs, we ran an experiment on the validation set, selecting documents with fewer than 200 pairs and considering only the BEFORE and SIMULTANEOUS relations. We applied the four-relations instruction (before, after, equal, and vague in our setting). The results are shown in Table~\ref{tab:maven_res}, and we observe that the GlobalConsistency method achieves results comparable to those reported on the four-relation news datasets, further confirming the method’s transferability across domains.

\section{Further Main Result Table Details}
\label{appx:main-res-details}
In this section we provide further details on the results and measurements presented in Table~\ref{tab:models-perform}

\subsection{Further Details on Reported Results}
We provide further details on the results presented in Table~\ref{tab:models-perform}. For the supervised models—RoBERTa, Bayesian, and Bayesian + Constraints—we report the best results achieved following a hyperparameter search (further detailed in Appendix~\ref{appx:experiment:details}). 
For the CoT experiment, we conducted a single evaluation run for each dataset and used this result. Constructing an ensemble or computing the mean for this experiment across multiple runs was beyond our budget.
In Table~\ref{tab:results-std}, we report the results for ZSL-Global and ZSL-Timeline, presenting the mean result obtained from five generations along with the standard deviation. For SelfConsistency and GlobalConsistency, we conducted a single run for each experiment, similar to CoT, as these experiments are more costly, and the observed standard deviation does not justify the additional expense.

\subsection{Transitive Inconsistency (TI) Details}
\label{appx:trans-inconsist}
For the Transitive Inconsistency (TI) measure reported in the table, we compute the average number of transitive-inconsistent edges per test document. We adopt a standard transitive closure algorithm \cite{warsheall-1962}, which is typically used to construct transitive relations. In our case, for any inferred path that implies a transitive relation, we verify whether the resulting relation is among the set of transitively allowed relations, as defined in \cite{ALLEN1984123, ning-etal-2018-joint} and related work. If the inferred relation violates this constraint, it is counted as a transitive inconsistency.

\subsection{Further Details on Event Count}
\label{appx:fig-expl:fig2}
In Figure~\ref{fig:cs-sent-expr-ds}, TB-Dense performance drops sharply for documents with over 25 events. Further analysis reveals that these documents predominantly contain \textit{vague} relations—considered more challenging and often associated with annotator disagreement \cite{chambers-etal-2014-dense}. ZSL-Timeline struggles with these relations (Figure~\ref{fig:per-rel-expr-ds}), particularly in TB-Dense and MATRES. As the frequency of \textit{vague} relations decreases beyond this threshold, performance improves.

\begin{table}[!t]
    \centering
    \resizebox{\columnwidth}{!}{
    \begin{tabular}{@{}l|c|c||c|c@{}}
        \toprule
        \makecell{\textbf{Model}} & \makecell{MATRES} & \makecell{TB-Dense} & \makecell{NT-6} & \makecell{\App{}} \\
        \midrule
        ZSL-Global (Ours) & 59.0±1.4 & 37.7±1.8 & 48.4±2.5 & 62.3±0.5 \\
        ZSL-Timeline (Ours) & 58.4±2.4 & 39.1±0.7 & 52.2±2.8 & 68.5±1.0 \\
        \bottomrule
    \end{tabular}}
    \caption{F1 scores of ZSL-Global and ZSL-Timeline are reported along with the standard deviation.}
    \label{tab:results-std}
\end{table}

\begin{table}[!t]
    \centering
    \small
    % \resizebox{0.35\textwidth}{!}{
    \begin{tabular}{@{}l|rrr@{}}
        \toprule
        & \makecell{Train} & \makecell{Dev} & \makecell{Test} \\
        \midrule
        MATRES & 13,577 & NA & 837 \\
        TB-Dense & 4,205 & 649 & 1,451 \\
        NarrativeTime & 68,317 & 2,759 & 7,925 \\
        \bottomrule
    \end{tabular}
    \caption{Statistics of event-event relations in the datasets used in this study.}
    \label{tab:dataset_all}
\end{table}

\section{Label Inconsistency Evaluation}
\label{appx:consist-eval}
We describe the \textit{Label Inconsistency} experiment detailed in §\ref{section:results:quality}. MATRES \cite{ning-etal-2018-multi} and TB-Dense \cite{chambers-etal-2014-dense} annotate the same set of 35 documents but follow different annotation schemes. MATRES considers only event start times to determine temporal order, while TB-Dense accounts for event start times, end times, and durations.

To isolate this difference, we define the following ground truth for each relation: (1) If a pair is marked as \textit{vague} in MATRES, meaning the event start time is unclear, the same pair should also be \textit{vague} in TB-Dense since both the start time and duration are uncertain. (2) If a pair in TB-Dense is annotated as \textit{before}, \textit{after}, or \textit{equal} based on both start and end times, the corresponding MATRES annotation should reflect the same relation when considering only event start times. Figure~\ref{fig:venn-diag} presents our findings in terms of label consistency and inconsistency between the two datasets.

\section{\App{} Annotation Proccess}
\label{appx:annot-process}
For the annotation process of \App{} (detailed in §\ref{section:background:annot-process}), we hired three annotators (two males and one female), all non-expert native English speakers and either undergraduate or graduate students. We instruct annotators to follow the MATRES annotation guidelines, considering only ``actual'' events (e.g., \textit{they \underline{won} the game}). Events that are ``non-actual'', such as intentional, negated, recurring, conditional, or wishful (e.g., \textit{I wish they \underline{win} the game}), are excluded from annotation. Additionally, only the starting time of events is considered when establishing temporal relations.

The actual annotation was done on 30 news summaries, each containing approximately 500 words. The annotators used the EventFull annotation tool \cite{eirew2024eventfullcompleteconsistentevent}, with all events in each document already highlighted. These events were extracted using the event detection method proposed by \citet{cattan-etal-2021-cross-document}, which identifies all types of events (actual and non-actual) and extracts an average of 60 event mentions per document, forming the initial set of events. We follow the same annotation protocol as proposed in EventFull. First, the annotation process begins with the selection of 15 to 18 of the most salient ``actual'' events from each story, following  \citet{eirew2024eventfullcompleteconsistentevent} which found that beyond 18 events, annotation becomes challenging for non-expert annotators. This event reduction aligns with previous efforts to decrease annotation workload by limiting the number of events considered \cite{chambers-etal-2014-dense, ning-etal-2018-multi, tan-etal-2024-set}. After selecting these events, each document was annotated for temporal relations (\textit{before}, \textit{after}, \textit{equal}, or \textit{vague}) by all three annotators. Finally, majority voting was used to determine the final relation, and in cases of disagreement, the relation was labeled as \textit{vague}.

Finally, the total annotation time for \App{}, including onboarding, amounted to 85 hours, with each worker paid $\$15$ per hour (which is considered a fair market value in their region). %Each document was annotated independently by all three annotators using the EventFull annotation tool.

% \section{Filling Transitive Relations}
% \label{appx:trans-rels}
% As discussed in §\ref{section:dataset:statistics}, to assess the coverage achievable by inferring transitive relations in resources annotated only with consecutive sentences, we extracted from NarrativeTime only the relations between event pairs in consecutive sentences. We then applied a transitive closure algorithm \cite{warsheall-1962} to construct additional relations and compared the results with the original set of relations. Figure~\ref{fig:nt_sentdiff} presents the experimental results.

\section{Formal Description of GlobalConsistency}
\label{appx:formal-zsl-global}
% ZSL-GlobalConsistency is formulated as follows: we run the ZSL-Timeline method five times on each input as described in §\ref{section:model}, generating five temporal graphs per document, denoted as  
% \(
% G = \{g_1, \dots, g_5\}
% \)
% where each \( g_n \) represents a labeled directed graph parsed from the DOT-language output. Each graph consists of a set of predicted event-pair relations:  
% \(
% g_n = \{p_{12}, p_{13}, \dots, p_{23}, p_{24}, \dots, p_{nm}\}
% \)
% where each relation \( p_{ij} \) is represented as a one-hot vector over the six relation types. We then sum these vectors element-wise across all five graphs and normalize them to obtain a single distribution per event pair:  
% \(
% d_{ij} = \frac{1}{5} \sum_{n=1}^{5} p_{ij}^{(n)}
% \)
% where each \( d_{ij} \) represents the normalized label distribution for the event pair \( (e_i, e_j) \). Instead of selecting the most frequent relation via majority voting, we apply the transitive constraints optimization algorithm, which returns a temporally consistent graph. We call this final method ZSL-GlobalConsistency (Figure~\ref{fig:figure1}). 

GlobalConsistency is formulated as follows: we run the ZSL-Timeline method five times on each input as described in §\ref{section:model}, generating five temporal graphs per document, denoted as  
\(
G = \{g_1, \dots, g_5\}
\)
where each \( g_n \) represents a labeled directed graph parsed from the DOT-language output. Each graph consists of a set of predicted event-pair relations:  
\(
g_n = \{p_{12}, p_{13}, \dots, p_{23}, p_{24}, \dots, p_{nm}\}
\)
where each relation \( p_{ij} \) is represented as a one-hot vector over the six relation types. We then sum these vectors element-wise across all five graphs and normalize them to obtain a single distribution per event pair:  
\(
d_{ij} = \frac{1}{5} \sum_{n=1}^{5} p_{ij}^{(n)}
\)
where each \( d_{ij} \) represents the normalized label distribution for the event pair \( (e_i, e_j) \). Instead of selecting the most frequent relation via majority voting, we apply a temporal constraints optimization algorithm, which returns a temporally consistent graph. We call this final method GlobalConsistency (Figure~\ref{fig:figure1}).

To perform this optimization, we define a binary decision variable \( \mathcal{I}_r(i, j) \in \{0, 1\} \) for each relation \( r \in \mathcal{R} \) and event pair \( (e_i, e_j) \), where \( \mathcal{R} \) is the set of possible temporal relations. The ILP objective is to maximize agreement with the model's predicted distributions:

\[
\max_{\mathcal{I}} \sum_{i \ne j} \sum_{r \in \mathcal{R}} \mathcal{I}_r(i, j) \cdot d_{ij}^r
\]

subject to the following constraints:

\begin{itemize}
    \item \textbf{Uniqueness:} Each event pair must be assigned exactly one relation:
    \[
    \sum_{r \in \mathcal{R}} \mathcal{I}_r(i, j) = 1 \quad \forall i \ne j
    \]
    
    \item \textbf{Symmetry:} For all inverse relations \( r \) and \( r^{-1} \), we ensure consistent labeling for reverse pairs:
    \[
    \mathcal{I}_r(i, j) = \mathcal{I}_{r^{-1}}(j, i) \quad \forall r \in \mathcal{R}
    \]
    
    \item \textbf{Transitivity:} For all event triplets \( (e_i, e_j, e_k) \), if \( \mathcal{I}_r(i, j) = 1 \) and \( \mathcal{I}_s(j, k) = 1 \), then \( \mathcal{I}_t(i, k) = 1 \) for some \( t \in \mathcal{C}(r, s) \), where \( \mathcal{C}(r, s) \subseteq \mathcal{R} \) defines the transitive closure over \( r \) and \( s \), as specified in \cite{ning-etal-2018-joint}.
\end{itemize}

This optimization ensures that the final output graph is both globally coherent and aligned with the model’s confidence across multiple generations.

\section{Dataset Licenses and Sources}
\label{append:dataset-license}
In our experiments, we use the following commonly used datasets for evaluating the temporal relation extraction task: MATRES \cite{ning-etal-2018-multi}, provided without a license; TimeBank-Dense \cite{chambers-etal-2014-dense}, provided without a license; and NarrativeTime \cite{rogers-etal-2024-narrativetime}, provided under the MIT license. 
Additionally, \App{} uses summaries from the Multi-News corpus \cite{fabbri-etal-2019-multi}, which is distributed under a custom license that permits free academic use.
All datasets were downloaded from official repositories, and used appropriately. \App{} will also be released under a free-to-use academic license.

\begin{figure}[t]
    \centering

    % First image with caption ONLY ABOVE
    \parbox{\linewidth}{ % Wrap in parbox so caption is above
        \centering{\footnotesize (a) \App{}.} \\ % Manually add caption above
        \includegraphics[width=0.98\linewidth]{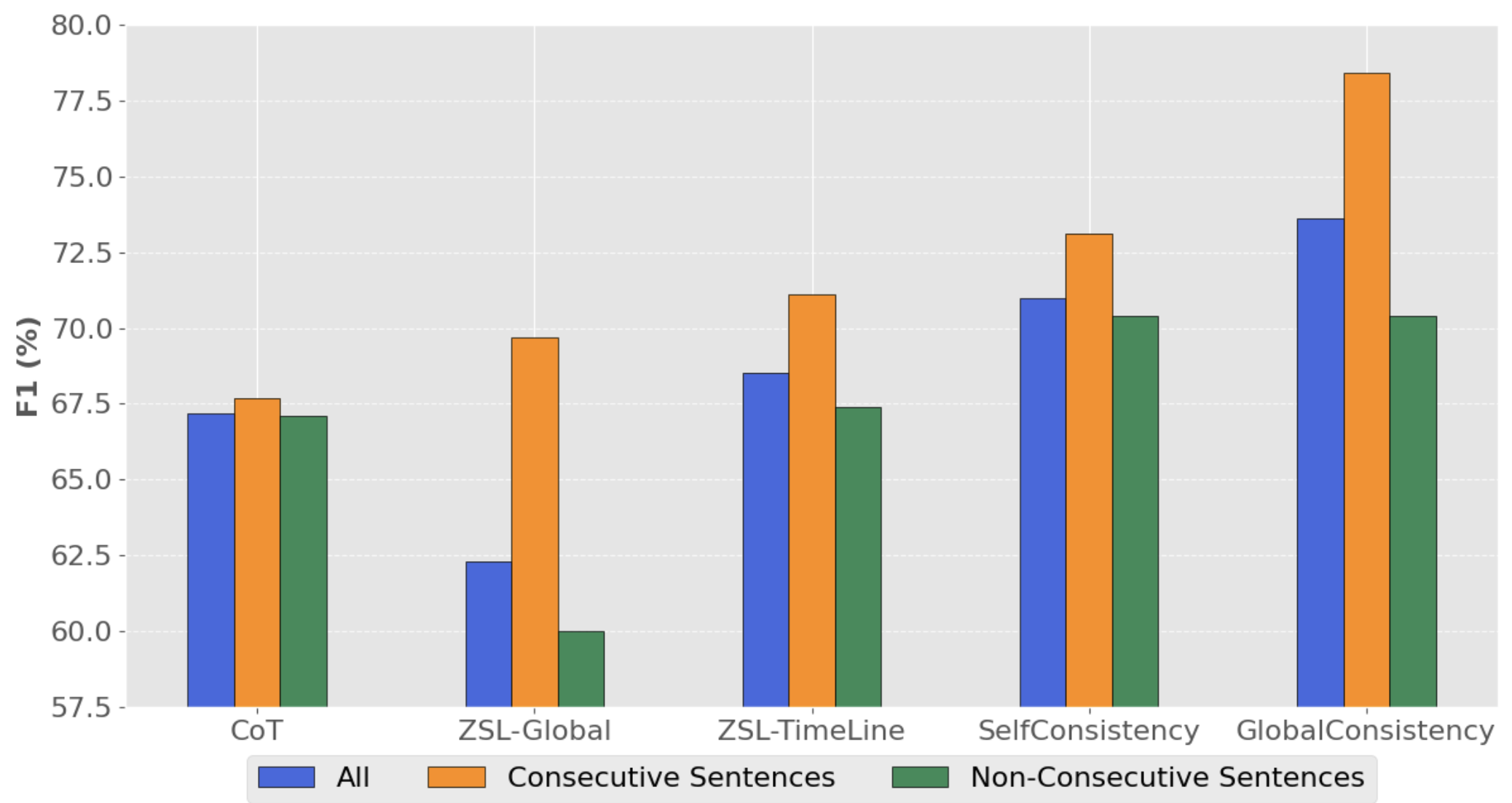}
    }

    \vspace{0.5cm} % Adjust spacing between figures

    % Second image with caption ONLY ABOVE
    \parbox{\linewidth}{
        \centering{\footnotesize (b) NarrativeTime.} \\ % Manually add caption above
        \includegraphics[width=0.98\linewidth]{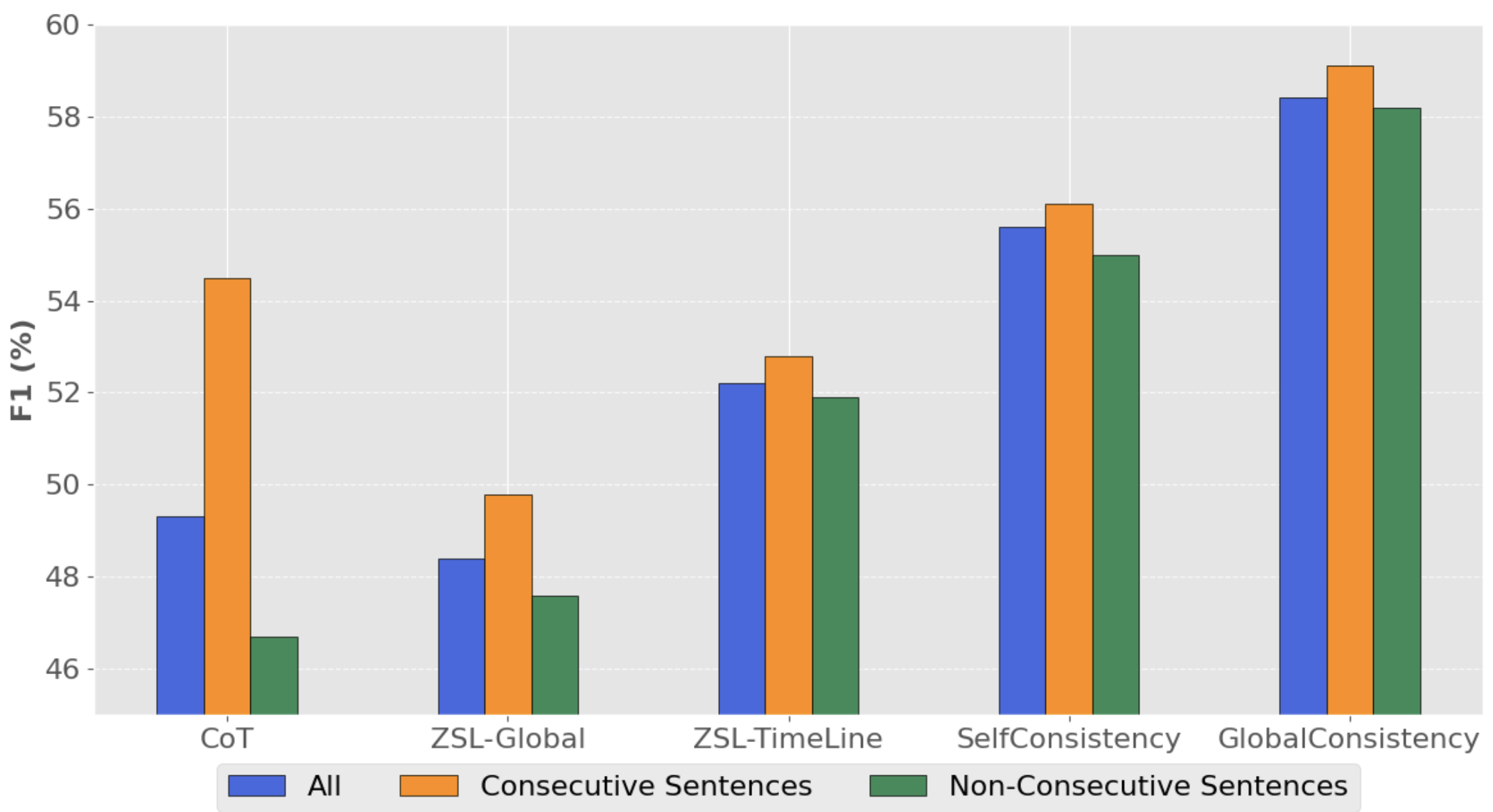}
    }

    % Main figure caption at the bottom
    \caption{Similar to Figure~\ref{fig:cs-sent-expr-ds}, we examine the performance across different relation subsets for GPT-4o.}
    \label{fig:cs-sent-expr-gpt}
\end{figure}

\begin{figure}[t]
    \centering

    % First image with caption ONLY ABOVE
    \parbox{0.95\linewidth}{ % Wrap in parbox so caption is above
        \centering{\footnotesize (a) NarrativeTime Vs. TimeBank-Dense.} \\ % Manually add caption above
        \includegraphics[width=0.98\linewidth]{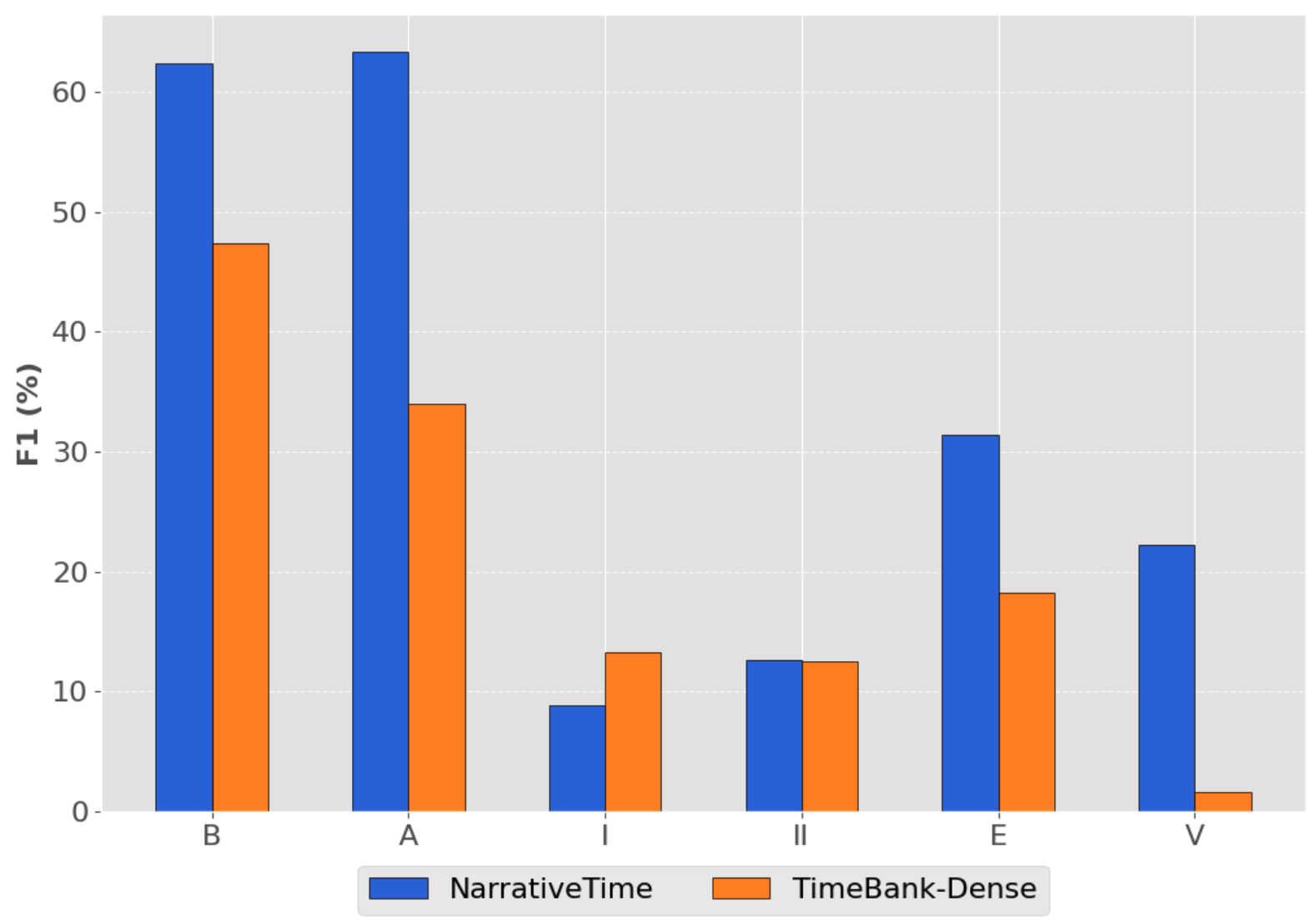}
    }

    \vspace{0.5cm} % Adjust spacing between figures

    % Second image with caption ONLY ABOVE
    \parbox{0.95\linewidth}{
        \centering{\footnotesize (b) \App{} Vs. MATRES.} \\ % Manually add caption above
        \includegraphics[width=0.98\linewidth]{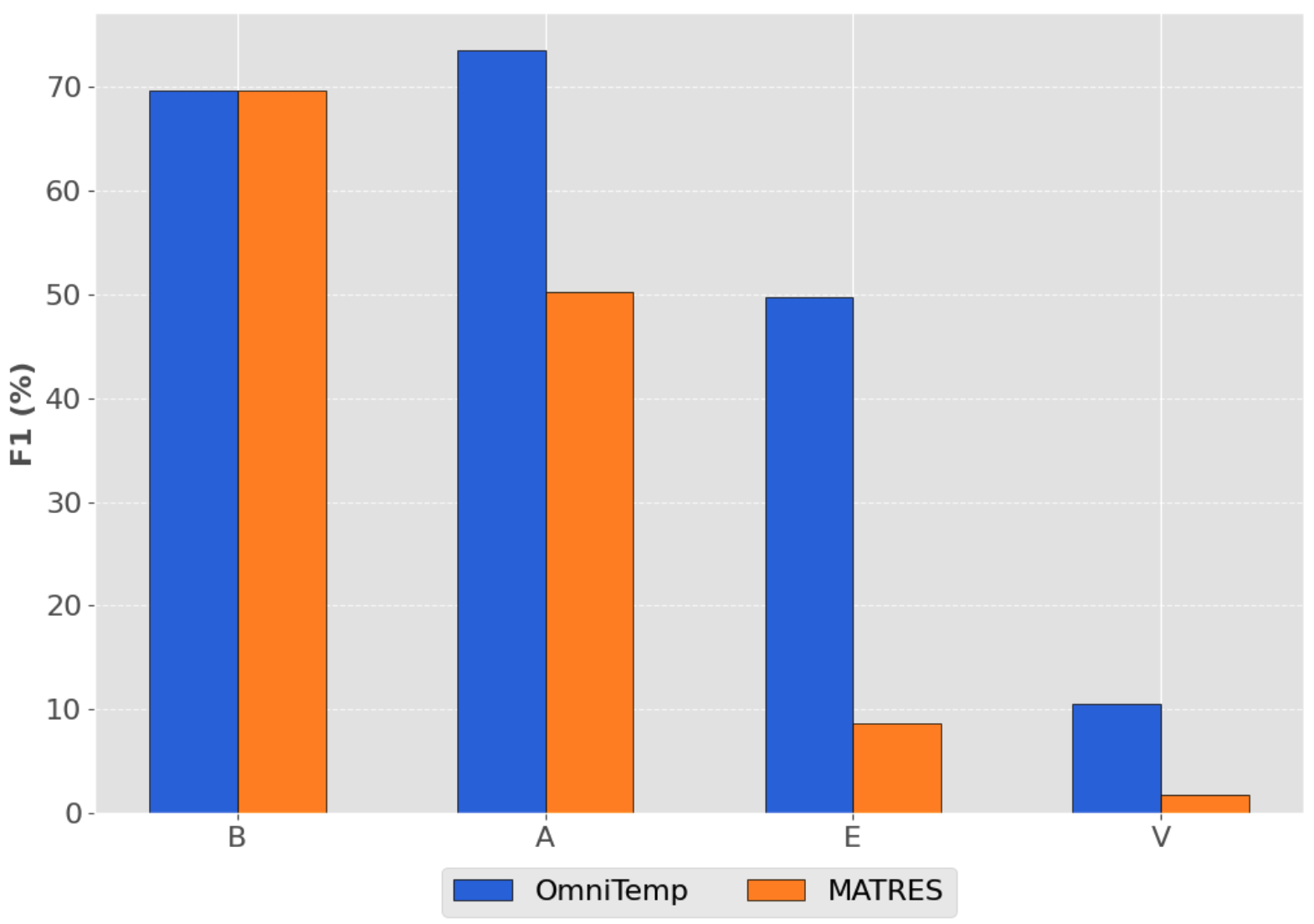}
    }

    % Main figure caption at the bottom
    \caption{Similar to Figure~\ref{fig:per-rel-expr-ds}, we examine the performance of our prompting method (i.e., ZSL-Timeline) by relation type using GPT-4o.}
    \label{fig:per-rel-expr-gpt}
\end{figure}

\section{Adjustments to the NarrativeTime Dataset}
\label{append:nt-further-details}
The NarrativeTime (NT) dataset, introduced in §\ref{section:background:datasets}, features seven relation types, including the six from TB-Dense and the \textit{overlap} relation. Our temporal consistency algorithm relies on Allen’s transitivity laws \cite{ALLEN1984123}, which require each relation type to have a symmetric counterpart (e.g., if event \textit{A} occurs \textit{before} event \textit{B}, then \textit{B} must occur \textit{after} \textit{A}). However, the \textit{overlap} relation in NT lacks a symmetric counterpart, making it incompatible for temporal consistency methods. 
%Additionally, the number of \textit{overlap} relations in the dataset is relatively small, further justifying its exclusion. 
Therefore, before using NT, we exclude event pairs labeled with the \textit{overlap} relation.
% The NarrativeTime (NT) dataset \cite{rogers-etal-2024-narrativetime} is already discussed in §\ref{section:background:datasets}. Before using NT, we exclude event pairs annotated with the \textit{overlap} relation, as its symmetric counterpart is missing from the dataset. This inconsistency conflicts with transitive constraint rules \cite{ALLEN1984123, ning-etal-2018-joint} and complicates our evaluation of temporal consistency (§\ref{section:model:pipeline}).  
Additionally, NT documents contain an average of 50 event mentions per document, corresponding to approximately 1,100 relations, which makes them difficult to process with LLMs due to context length limitations. Handling such documents requires segmenting them and making individual calls to the model for each segment, which increases costs, as discussed in §\ref{section:model}. To avoid segmentation and reduce costs, we randomly select 18 events per document from the test set, along with all their associated relations. The choice of 18 events was based on empirical observations, as it represents the maximum number that can typically fit within the model's context window without requiring segmentation. This reduction is not applied to the training set, which we use to fine-tune the supervised models. We refer to this pre-processed version as NT-6, as it retains only six relation types.

\section{Additional Experiment Tables and Figures}
\label{append:additional-figures}
Figure~\ref{fig:per-rel-expr-gpt} presents the relation-wise performance of GPT-4o, analogous to the results shown for DeepSeek-R1 in Figure~\ref{fig:per-rel-expr-ds}. Figure~\ref{fig:cs-sent-expr-gpt} presents the model performance across different relation subsets, analogous to the results shown for DeepSeek-R1 in Figure~\ref{fig:cs-sent-expr-ds}. Table~\ref{tab:stats_all} presents a comparison between common datasets used for evaluating models on the temporal relation task alongside \App{}. Table~\ref{tab:dataset_all} presents the split statistics of these datasets. Figure~\ref{fig:zsl-global-prompt} presents an example of the ZSL-Global prompt. Figure~\ref{fig:timeline-output} presents an example of the generated timeline using the ZSL-Timeline approach. Figure~\ref{fig:nt_sentdiff} presents the experimental results for filling transitive relations in a dataset containing only temporal relations between events up to one sentence apart (similar to MATRES and TB-Dense).

\begin{figure*}[t]
    \centering
    \subfloat[Before]{\includegraphics[width=0.40\linewidth]{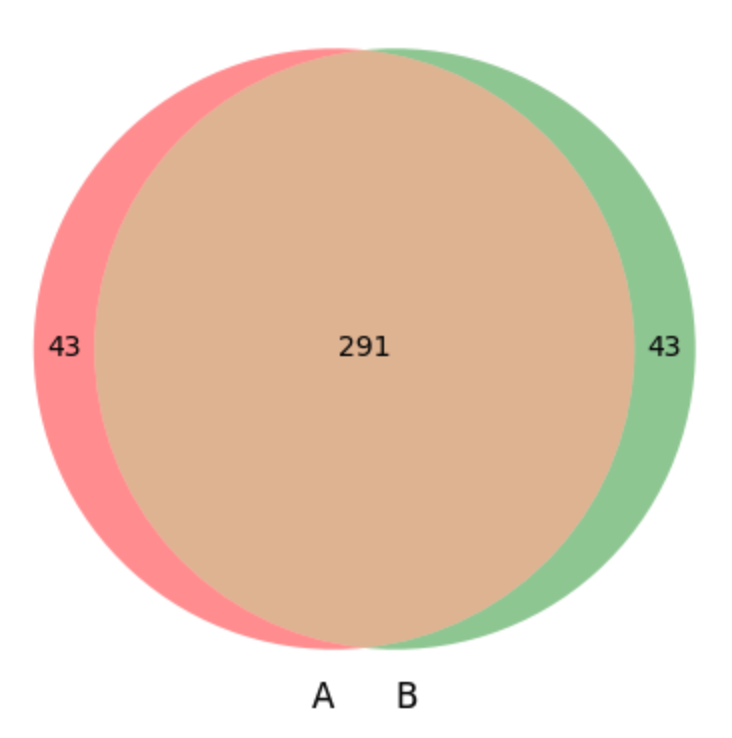}\label{fig:venn:first}}
    \hfill
    \subfloat[After]{\includegraphics[width=0.40\linewidth]{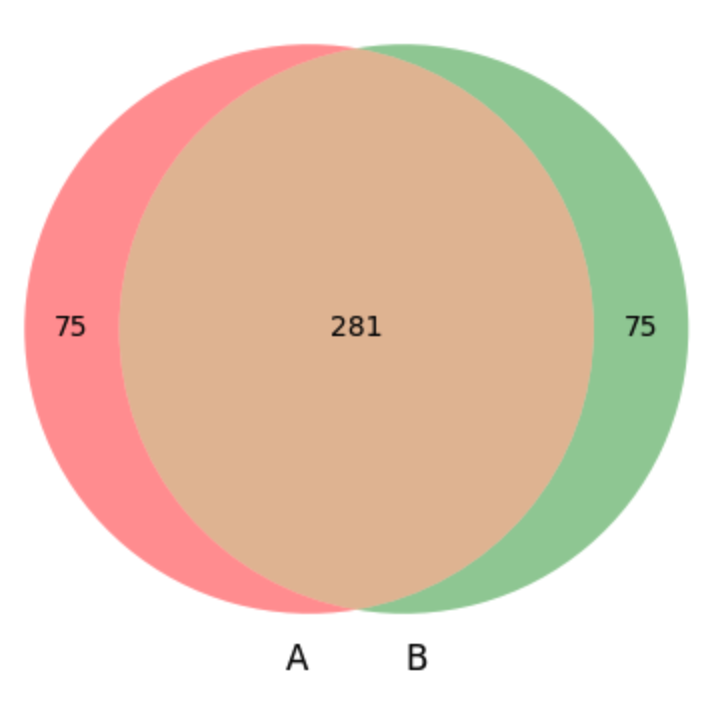}}
    \hfill
    \subfloat[Vague]{\includegraphics[width=0.40\linewidth]{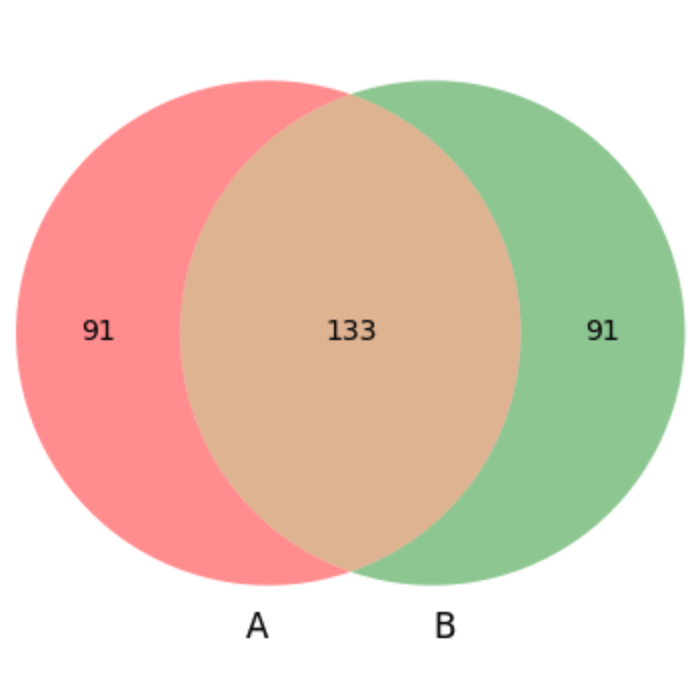}}\label{fig:venn:third}
    \hfill
    \subfloat[Equal]{\includegraphics[width=0.40\linewidth]{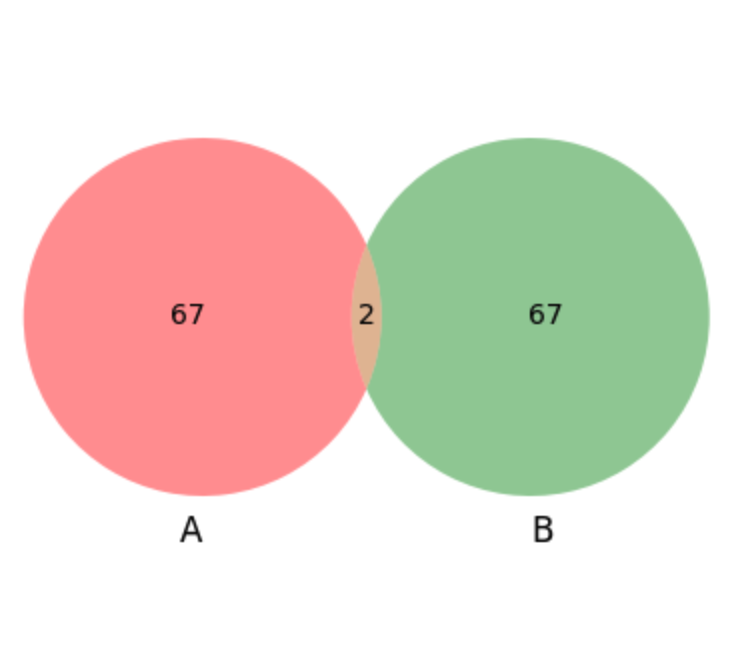}}
    \label{fig:venn:fourth}
    \caption{Label Inconsistency: Each group, A and B, represents MATRES and TimeBank-Dense respectively. The intersecting area indicates consistency in label annotation between the two datasets, with the number of such pairs highlighted in the middle, while the non-intersecting areas represent pairs assigned different labels in each dataset.}
    \label{fig:venn-diag}
\end{figure*}

\begin{figure*}[t!]
\centering
\includegraphics[width=0.7\textwidth]{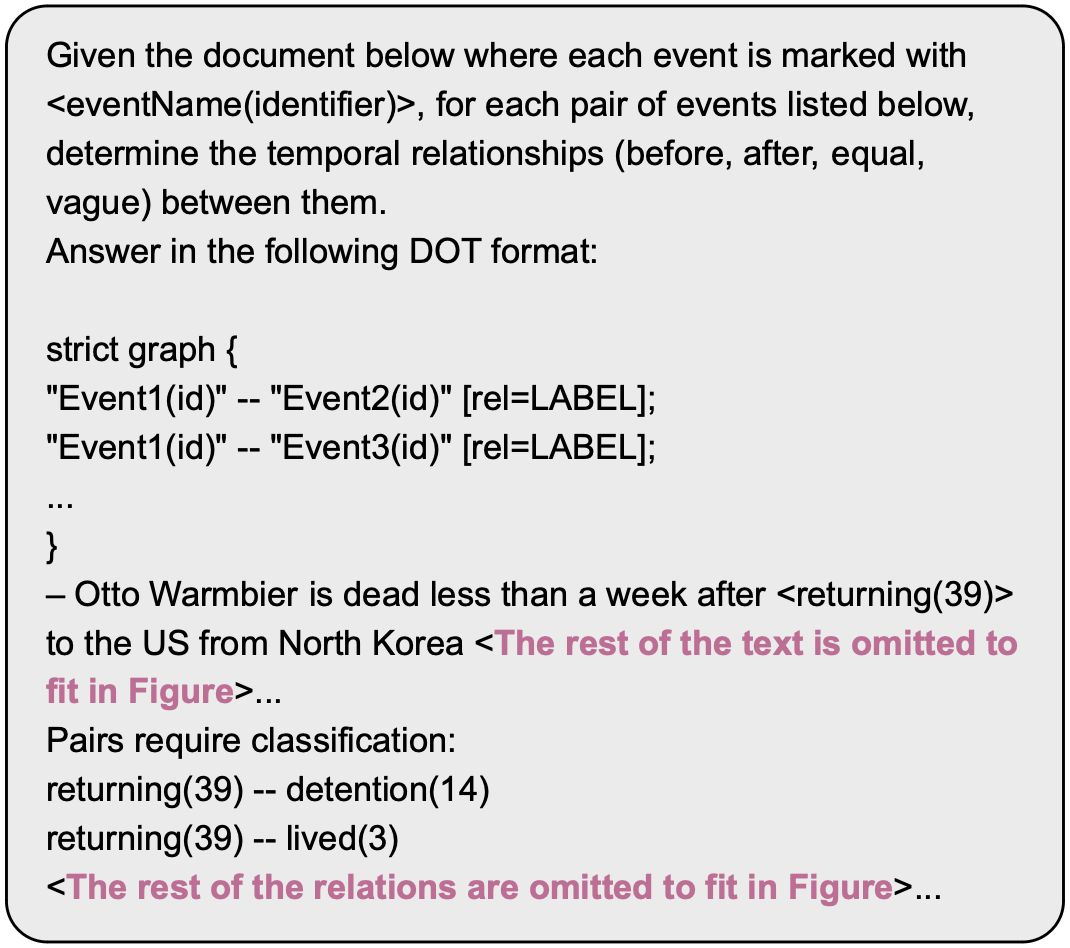}
\caption{An example of the ZSL-Global prompt.}
\label{fig:zsl-global-prompt}
\end{figure*}

\begin{figure*}[t!]
\centering
\includegraphics[width=\textwidth]{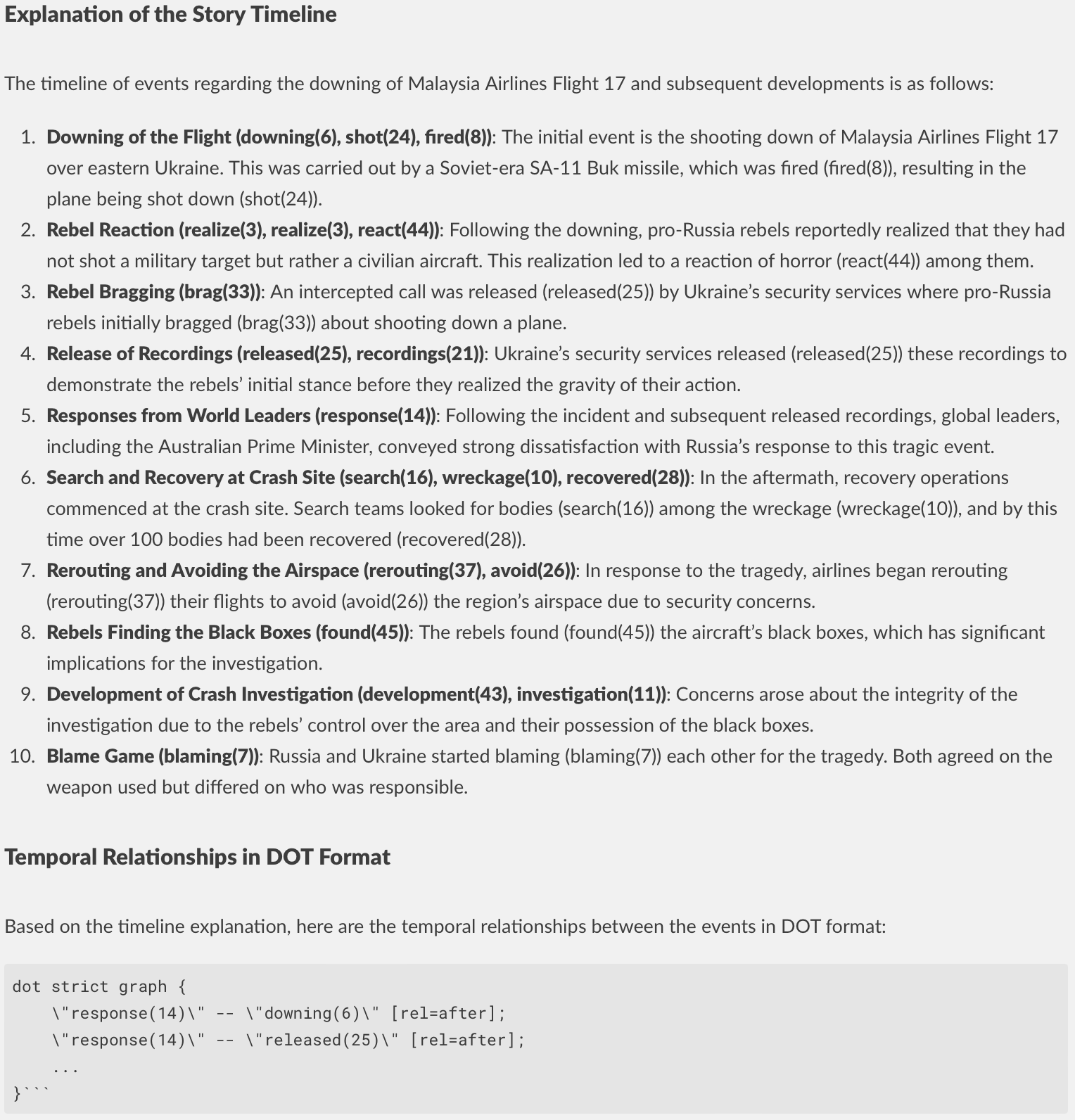}
\caption{An example of a generated output when GPT-4o is prompted using the ZSL-Timeline method (with the Markdown format retained from the original output). The full event list is generated; however, it is trimmed (indicated by ``...'') in this example to ensure the output fits within the figure.}
\label{fig:timeline-output}
\end{figure*}

\begin{table*}[!t]
    \centering
    \resizebox{0.8\textwidth}{!}{
    \begin{tabular}{@{}l|c|c|c|c|c||c@{}}
        \toprule
        & \makecell{MATRES} & \makecell{TB-Dense} & \makecell{TCR} & \makecell{TDD-Manual} & \makecell{NarrativeTime} & \makecell{\textbf{\App{}}} \\
        \midrule
        \multicolumn{7}{c}{\textbf{Datasets Statistics}} \\
        \midrule
        Documents & 275 & 36 & 25 & 34 & 36 & 30 \\
        Events & 6,099 & 1,498 & 1,134 & 1,101 & 1,715 & 470 \\
        \midrule
        \textit{before} & 6,852 (50) & 1,361 (21) & 1,780 (67) & 1,561 (25) & 17,011 (22) & 1,540 (44) \\
        \textit{after} & 4,752 (35) & 1,182 (19) & 862 (33) & 1,054 (17) & 18,366 (23) & 1,347 (39) \\
        \textit{equal} & 448 (4) & 237 (4) & 4 (0) & 140 (2) & 5,298 (7) & 150 (4) \\
        \textit{vague} & 1,525 (11) & 2,837 (45) & -- & -- & 25,679 (33) & 446 (13) \\
        \textit{includes} & -- & 305 (5) & -- & 2,008 (33) & 5,781 (7) & -- \\
        \textit{is-included} & -- & 383 (6) & -- & 1,387 (23) & 6,639 (8) & -- \\
        \textit{overlaps} & -- & -- & -- & -- & 227 (0) & -- \\
        \midrule
        Total Relations & 13,577 & 6,305 & 2,646 & 6,150 & 79,001 & 3,483 \\
        \midrule
        \multicolumn{7}{c}{\textbf{Per Document Average Annotation Sparsity}} \\
        \midrule
        Events & 22.2 & 41.6 & 45.4 & 32.4 & 47.6 & 15.6 \\
        Actual Relations & 49.4 & 183.7 & 105.8 & 180.9 & 1,110.1 & 114.9 \\
        Expected Relations & 234.8 & 844.5 & 1,006.1 & 508.1 & 1,110.1 & 114.9 \\
        \midrule
        Missing Relations & 79\% & 78.3\% & 89.5\% & 64.4\% & 0\% & 0\% \\
        \bottomrule
    \end{tabular}}
    \caption{The upper part of the table presents the statistics of notable datasets for the temporal relation extraction task alongside \App{}. In parentheses, the values indicate the percentage of each relation type relative to the total relations in the dataset. The bottom part of the table summarizes the average percentage of missing relations per document, calculated as the ratio of actual annotated relations to a complete relation coverage, referred to as \textit{Expected Relations}.}
    \label{tab:stats_all}
\end{table*}

% \begin{table*}[!t]
%     \centering
%     \resizebox{0.8\textwidth}{!}{
%     \begin{tabular}{@{}l|c|c|c|c|c|c@{}}
%         \toprule
%         & \makecell{MATRES} & \makecell{TBD} & \makecell{TCR} & \makecell{TDD-Man} & \makecell{NarrativeTime} & \makecell{\App{}} \\
%         \midrule
%         Docs & 275 & 36 & 25 & 34 & 36 & 30 \\
%         Events & 6,099 & 1,498 & 1,134 & 1,101 & 1,715 & 470 \\
%         \midrule
%         Before (\%) & 6,852 (50) & 1,361 (21) & 1,780 (67) & 1,561 (25) & 17,011 (22) & 1,540 (44) \\
%         After (\%) & 4,752 (35) & 1,182 (19) & 862 (33) & 1,054 (17) & 18,366 (23) & 1,347 (39) \\
%         Equal (\%) & 448 (4) & 237 (4) & 4 (0) & 140 (2) & 5,298 (7) & 150 (4) \\
%         Vague (\%) & 1,525 (11) & 2,837 (45) & -- & -- & 25,679 (33) & 446 (13) \\
%         Includes (\%) & -- & 305 (5) & -- & 2,008 (33) & 5,781 (7) & -- \\
%         IsIncluded (\%) & -- & 383 (6) & -- & 1,387 (23) & 6,639 (8) & -- \\
%         Overlaps (\%) & -- & -- & -- & -- & 227 (0) & -- \\
%         \midrule
%         Total Rels & 13,577 & 6,305 & 2,646 & 6,150 & 79,001 & 3,483 \\
%         \bottomrule
%     \end{tabular}}
%     \caption{Statistics of notable datasets for the temporal relation extraction task.}
%     \label{tab:stats}
% \end{table*}

\begin{figure*}[t!]
\centering
\includegraphics[width=\textwidth]{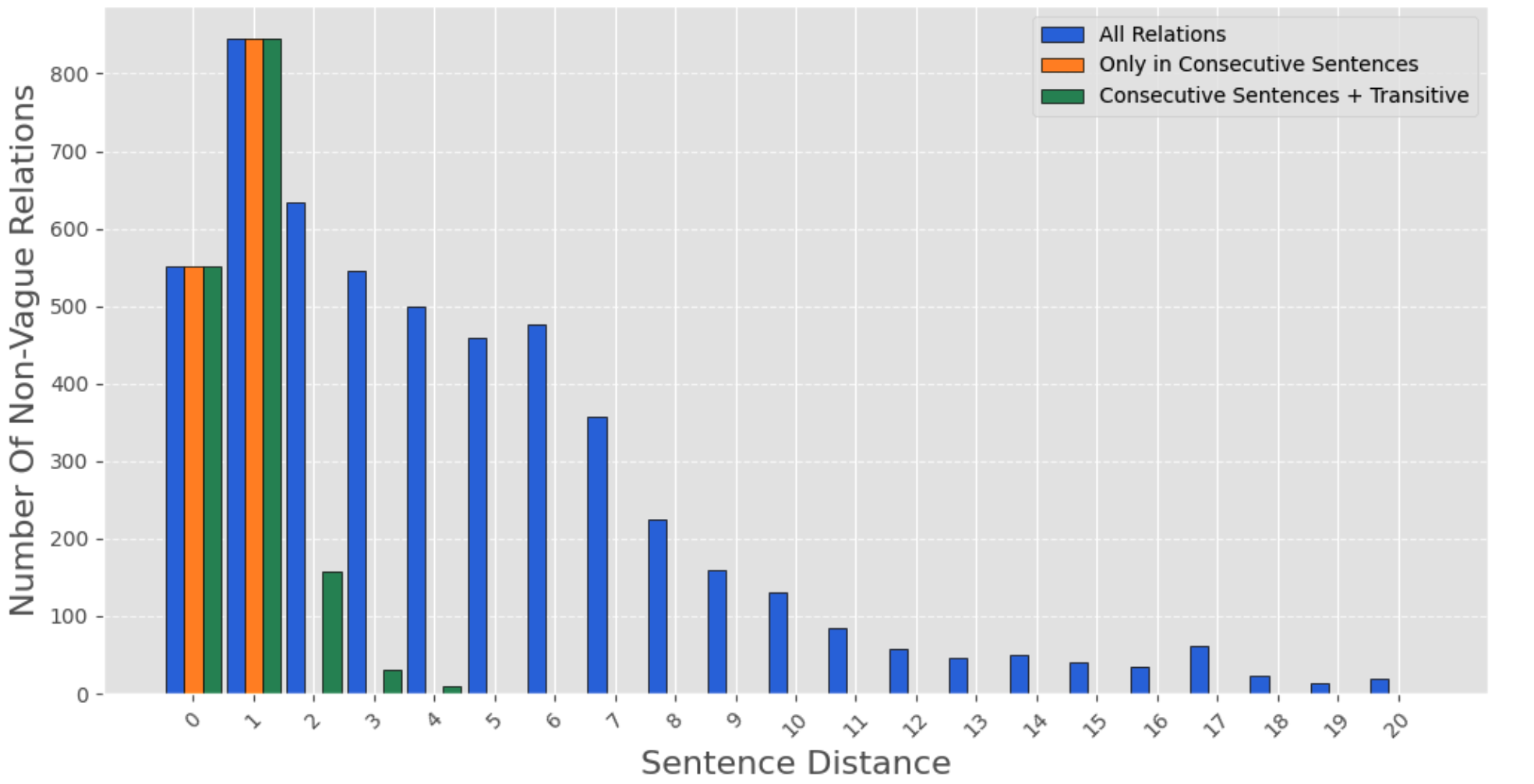}
\caption{Illustration of the achieved relation distance after applying transitive closure in resources annotated only between consecutive sentences. The blue bars represent the original set of relations in NarrativeTime, which is exhaustively annotated between all events. The orange bars represent the version created by considering only relations between events in consecutive sentences. The green bars represent the set of relations after applying a transitive algorithm to infer additional relations.}
\label{fig:nt_sentdiff}
\end{figure*}

\end{document}